\documentclass[onecolumn,10pt]{IEEEtran}

\usepackage{amsfonts, amssymb, amsmath}
\usepackage{epsfig}
\usepackage{theorem}
\usepackage{textcomp}
\usepackage{times}
\usepackage{multicol}
\usepackage{multirow}
\usepackage{tabularx}
\usepackage{graphicx}
\usepackage[justification=centering]{caption}

\makeatletter
\let\NAT@parse\undefined
\makeatletter
\usepackage{natbib}

\def\mpp{{\sc Mpp}}

\makeatletter 
\newcommand\semiHuge{\@setfontsize\semiHuge{22.72}{27.38}}
\makeatother

\begin{document}
\title{An Effective Algorithmic Framework for Near Optimal Multi-Robot Path Planning}
\author{Jingjin Yu\qquad\qquad Daniela Rus
\thanks{Jingjin Yu and Deniela Rus are with the Computer Science and Artificial Intelligence Lab, Massachusetts Institute of Technology. E-mails: \{jingjin, rus\}@csail.mit.edu.}%
}
\date{}
\maketitle
\begin{abstract} We present a centralized algorithmic framework for solving  multi-robot path planning problems in general, two-dimensional, continuous environments while minimizing globally the task completion time. The framework obtains high levels of effectiveness through the composition of an optimal discretization of the continuous environment and the subsequent fast, near-optimal resolution of the resulting discrete planning problem. This principled approach achieves orders of magnitudes better performance with respect to both speed and the supported robot density. For a wide variety of environments, our method is shown to compute globally near-optimal solutions for $50$ robots in seconds with robots packed close to each other. In the extreme, the method can consistently solve problems with hundreds of robots that occupy over $30\%$ of the free space. 
\end{abstract}

\section{Introduction}\label{section:introduction}
\begin{figure}[ht!]
\begin{center}
  \begin{tabular}{ccc}
    \includegraphics[width=3in]{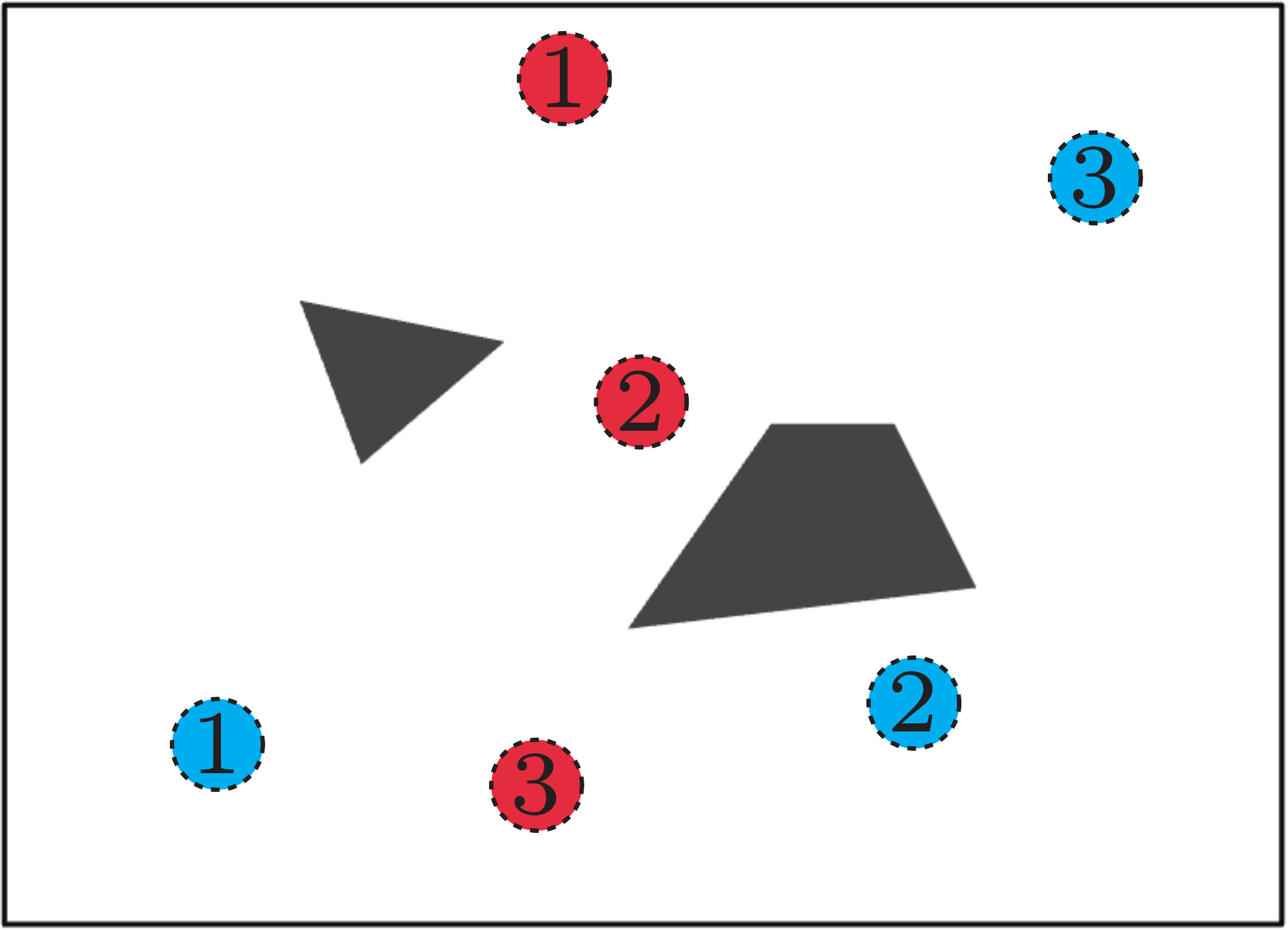} & \hspace{10mm} & 
		\includegraphics[width=3in]{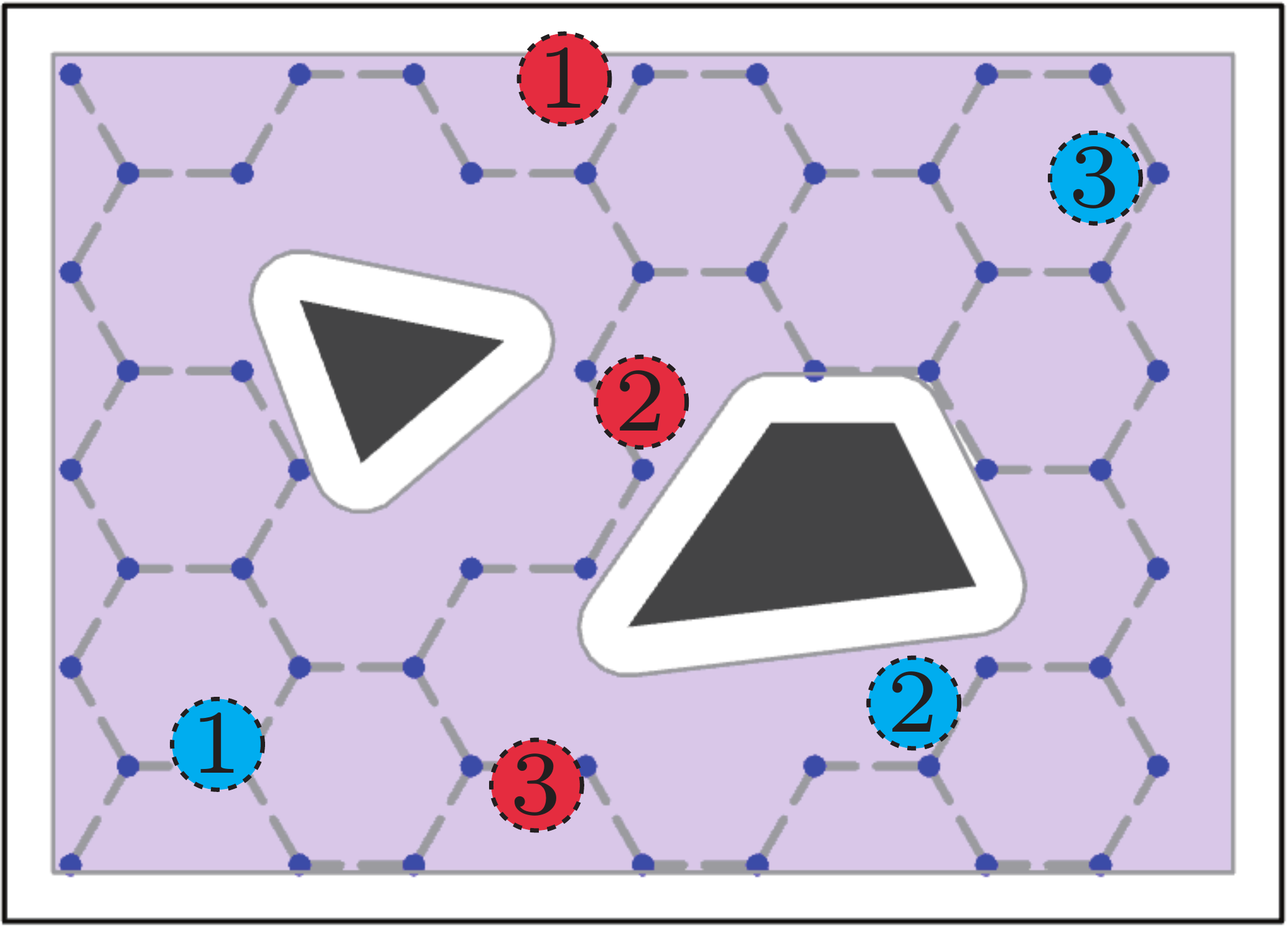}  \\ \vspace{1mm}
    (a) & & (b) \\ 
    \includegraphics[width=3in]{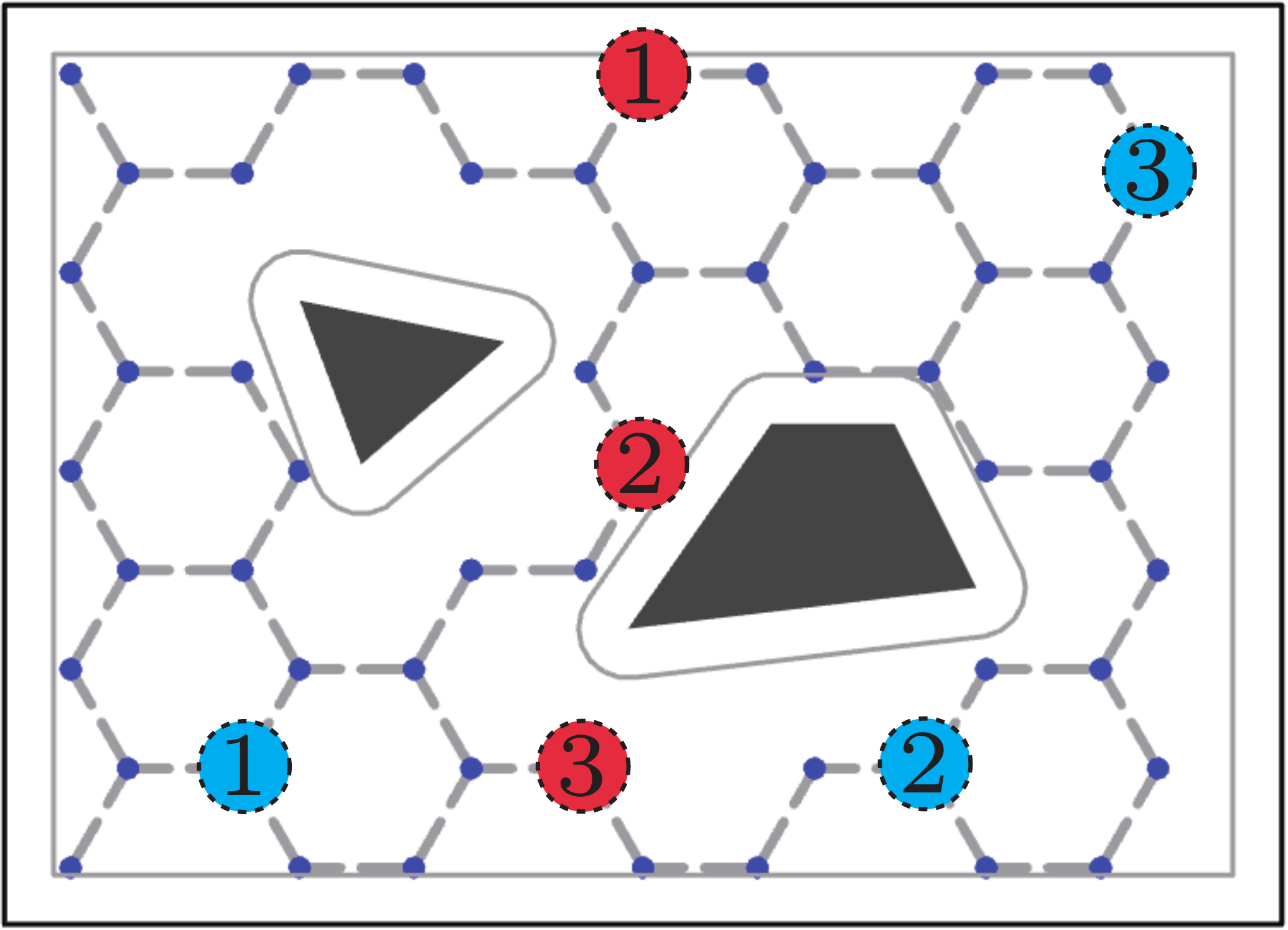} & &
		\includegraphics[width=3in]{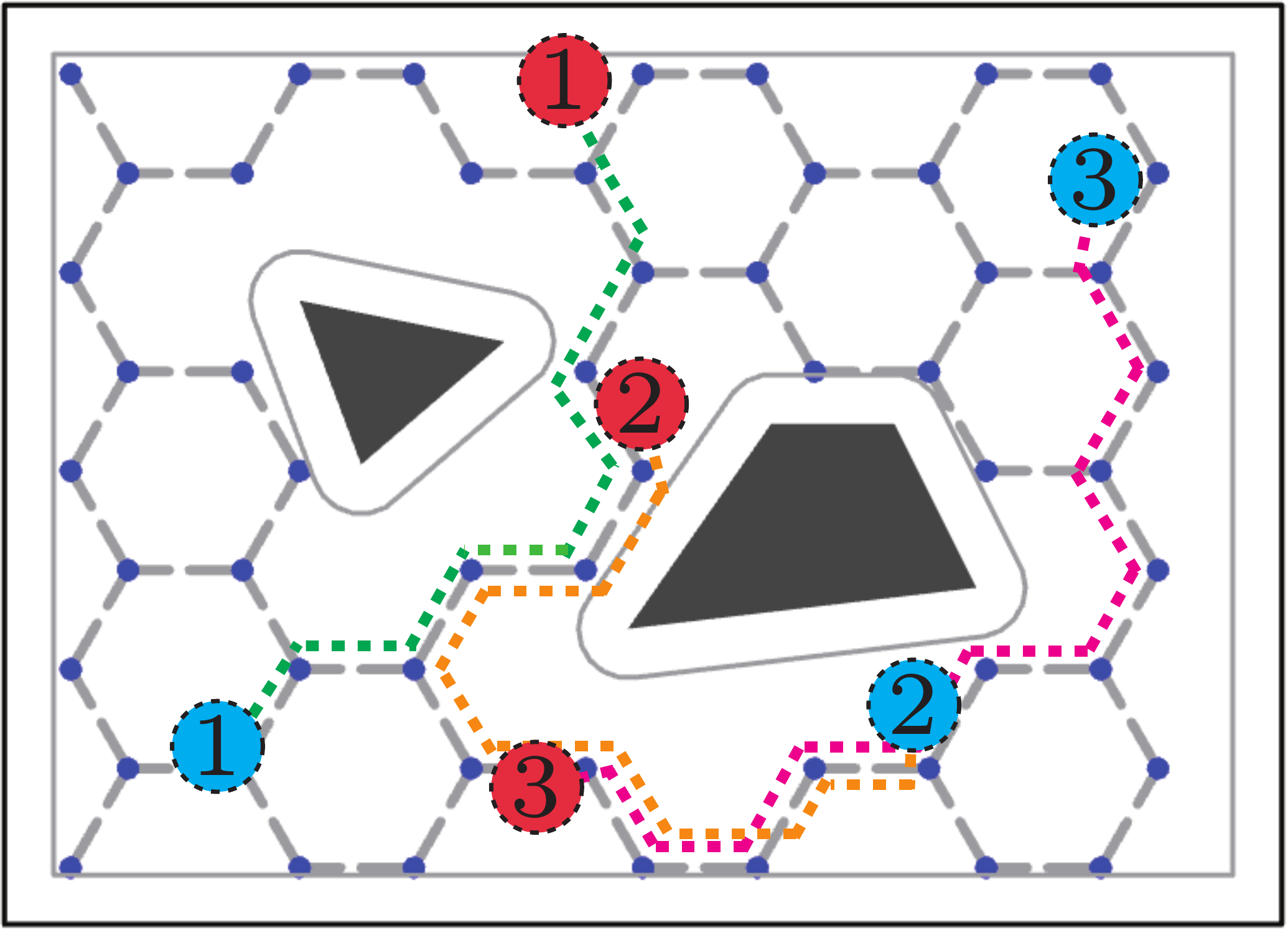}  \\ \vspace{1mm}
    (c) & &(d)\\
  \end{tabular}
\end{center}
\caption{\label{figure:example} An illustrative example of our algorithmic framework. a) A problem instance with three disc robots. The start and goal locations are indicated by the blue and red labeled discs, respectively. b) The configuration space (shaded area) for a single robot and the fitted hexagonal lattice. The blue circles are the start positions, and the red circles are the goal positions. c) The discrete abstraction of the original problem. d) Solution to the original continuous problem.}
\end{figure}

We study the problem of planning collision-free paths for multiple labeled disc robots operating in two-dimensional, multiply-connected, continuous environments ({\em i.e.}, environments with holes). The {\em primary goal} of this work is to develop a  practical, extensible framework toward the efficient resolution of multi-robot path planning (\mpp) problems, in which the robots are densely packed, while simultaneously seeking to minimize {\em globally the task completion time}. The framework is composed of two key algorithmic components, executed in an sequential order. Using the example illustrated in Fig.~\ref{figure:example}(a), first, we compute the configuration space for a single robot, over which an optimal lattice structure is overlaid (Fig.~\ref{figure:example}(b)). Using the lattice structure as a roadmap, each start (resp., goal) location is assigned to a nearby node of the roadmap as its unique discrete start (resp., goal) node, which translates the continuous problem into a discrete one (Fig.~\ref{figure:example}(c)). Then, a state-of-the-art discrete planning algorithm is applied to solve the roadmap-based problem near-optimally (Fig.~\ref{figure:example}(d)). Through the tight composition of these two algorithmic components, our framework proves to be highly effective in a variety of settings, pushing the boundaries on optimal multi-robot path planning to new grounds in terms of the number of robots supported and the allowed robot density. 

{\bf Related work}. \mpp\, finds applications in a wide spectrum of domains such as navigation \cite{AlonsoMora:2014id,SnaGuyBerMan14}, manufacturing and assembly \cite{knepper2012pedestrian}, warehouse automation \cite{WurDanMou08}, computer video games \cite{Sna12}, and microfluidics \cite{GriAke05}. Given the important role it holds in robotics-related applications, \mpp\, problems has received considerable attention in robotics research with dedicated study on the subject dating back at least three decades \cite{schwartz1983piano}, in which a centralized approach is taken that considers all robots as a single entity in a high dimensional configuration space. Because the search space in such problems grows exponentially as the number of robots increases linearly, a centralized approach \cite{schwartz1983piano}, although complete, would be extremely inefficient in practice. As such, most ensuing research take the approach of decomposing the problem. One way to do this is by assigning priorities to the robots so that robots with higher priority take precedence over robots with lower priority \cite{buckley1989fast,ErdLoz86}. Another often adopted partitioning method is to plan a path for each robot separately without considering robot-robot interaction. The paths are then {\em coordinated} to yield collision free paths \cite{bien1992minimum,OdoLoz89}. Following these initial efforts, the decomposition scheme is further exploited and improved \cite{GhrOkaLav05,LavHut98b,PenAke02,BerOve05,BerSnoLinMan09,SveOve98}. Many of the mentioned works also consider optimality in some form. We emphasize that, since finding feasible solution for \mpp\, is already PSPACE-hard \cite{hopcroft1984complexity}, {\em i.e.}, no polynomial-time complete algorithm may even exist for such problems unless P = PSPACE, computing globally near-optimal solution for a large number of robots is extremely challenging. 

Recent years have witnessed a great many new approaches being proposed for solving \mpp. One such method, reciprocal velocity obstacles \cite{blm-rvo,BerSnaGuyMan11}, which can be traced back to \cite{KanZuc86}, explicitly looks at velocity-time space for coordinating robot motions. In \cite{GriAke05}, mixed integer programming (MIP) models are employed to encode the interactions between the robots. A method based on network-flow is explored in \cite{KarGerSta12}. In \cite{PeaClaMcp08}, similar to our framework upon a first look, an $A^*$-based search is performed over a discrete roadmap abstracted from the continuous environment. However, the authors addressed a much narrower class of problems for which they can bound the computation cost but cannot guarantee the solution optimality. It is also unclear how the complex geometric problem of efficiently computing a discrete roadmap from the continuous environment is resolved in the paper. In \cite{SolSalHal14}, discrete-RRT (d-RRT) is proposed for the efficient search of multi-robot roadmaps. Lastly, as a special case of \mpp\, in continuous domains, efficient algorithms are proposed \cite{SolYu15,TurMicKum13ICRA} for interchangeable robots ({\em i.e.}, in the end, the only requirement is that each goal location is occupied by an arbitrary robot). At the same time, discrete ({\em e.g.}, graph-based) \mpp\, has also been a subject of active investigation. This line of research originates from the mathematical study of the 15-puzzle and related {\em pebble motion} problems \cite{KorMilSpi84,Wil74}. Since then, many heuristics augmenting the $A^*$ algorithm have been proposed for finding optimal solution, {\em e.g.}, \cite{Rya08,StaKor11,WagChoC11}, to name a few. These heuristics essentially explore the same decoupling idea used in the continuous case to trim down the search space. A method based on network-flow also exists here \cite{YuLav13ICRA-A}. Some of these discrete solutions, such as \cite{KorMilSpi84}, have helped solving continuous problems \cite{ksb-tdmp12,SolHal12}. 

{\bf Contribution.} Our work brings two contributions toward solving \mpp\, effectively and optimally. First, we introduce a two-phase framework that allows any roadmap building ({\em i.e.}, discretization) method to be combined with any suitable discrete \mpp\, algorithm for solving continuous \mpp\, problems. The framework achieves this by imposing a partial collision avoidance constraint during the roadmap building phase while preserving path near-optimality. Second, we deliver a practical integrated algorithmic implementation of the two-phase framework for computing near optimal paths for a large number of robots. We accomplish this by combining {\em (i)} a fast algorithm for superimposing dense regular lattice structures over a bounded two-dimensional environment with holes and {\em (ii)} an integer linear programming (ILP) based algorithm for computing near-time-optimal solutions to discrete \mpp\,\cite{YuLav13AAAI-LBP}. To the best of our knowledge, we present the first such algorithm that can quickly plan near optimal, continuous paths for hundreds of robots densely populated in multiply-connected environments\footnote{Warehousing systems from Kiva Systems \cite{WurDanMou08} can work effectively with hundreds of robots. However, these robots essentially live on a grid within a structured environment.}.

{\bf Paper organization.} The rest of the paper is organized as follows. We formulate the \mpp\,problem in Section \ref{section:formulation}. In Section~\ref{section:algorithm}, we describe the overall algorithmic framework architecture and the first component of the framework on roadmap-based problem construction. In Section~\ref{section:splitting}, we describe how the second component of the framework may be realized. In Section \ref{section:experiment}, we demonstrate the effectiveness of our framework over a variety of environments. We hold an extensive discussion and conclude in Section \ref{section:conclusion}.\footnote{An accompanying video demonstrating our algorithm and software developed in this paper are available from the corresponding author's website.}

\section{Problem Statement}\label{section:formulation}
Let $\mathcal W$ denote a bounded, open, multiply-connected ({\em i.e.}, with holes), two-dimensional region. We assume that the boundary and obstacles of $\mathcal W$ can be approximated using polygons with an overall complexity of $m$ ({\em i.e}, there are a total of $m$ edges). There are $n$ unit disc robots residing in $\mathcal W$. These robots are assumed be omnidirectional with a velocity $v$ satisfying $\vert v \vert \in [0, 1]$. Let $\mathcal C_f$ denote the free configuration space for a single robot (the shaded area in Fig. \ref{figure:example}(b)). The centers of the $n$ robots are initially located at $S = \{s_1, \ldots, s_n\} \subset \mathcal C_f$, with goals $G = \{g_1, \ldots, g_n\} \subset \mathcal C_f$. For all $1 \le i \le n$, a robot initially located at $s_i$ must be moved to $g_i$. 

In addition to planning collision-free paths, we are interested in optimizing path quality. Our particular focus in this paper is minimizing the {\em global task completion time}, also commonly known as {\em makespan}\footnote{Note that our algorithmic framework also applies to other time- and distance-based optimality objectives through the use of an appropriate discrete planning algorithm.}. Let $P = \{p_1, \ldots, p_n\}$ denote a feasible path set with each $p_i$ a continuous function, defined as 
\[
p_i: [0, t_f] \to C_f, p_i(0) = s_i, p_i(t_f) = g_i.
\]
The {\em makespan} objective seeks solutions that minimize $t_f$. In other words, let $\mathcal P$ denote the set of all solution path sets, the task is to find a path set with $t_f$ close to
\begin{align}\label{equation:minimum-time}
t_{min} := \min_{P \in \mathcal P} t_f(P). 
\end{align}

We emphasize that the aim of this work is a method for quickly solving ``typical'' problem instances with many robots and high robot density ({\em i.e.}, the ratio between robot footprint and the free space is high) with optimality assurance. By {\em typical}, we mean that: {\em (i)} the start location and goal locations are reasonably separated, {\em (ii)} a start or goal location is not too close to static obstacles in the environment, and {\em (iii)} there are no narrow passages in the environment that cause the discretized roadmap structure to have poor connectivity. More formally, we assume that assumptions {\em (i)} and {\em (ii)}, respectively, take the forms \footnote{\eqref{equation:separation} and ~\eqref{equation:obstacle-separation} are unit-less given the unit disc robot assumption. If the robots have radius $r$, the right side of the inequalities from~\eqref{equation:separation} and ~\eqref{equation:obstacle-separation} should be scaled by a multiplicative factor of $r$.}
\begin{align}
\forall 1 \le i, j \le n,\quad \vert s_i - s_j \vert \ge 4,\>\> \vert g_i - g_j \vert \ge 4 \label{equation:separation}
\end{align}
and 
\begin{align}
\forall p \in \{S \cup G\}, \quad \vert p - q \vert \le \sqrt{5} \Rightarrow q \in \mathcal W. \label{equation:obstacle-separation}
\end{align}

\begin{figure}[ht!]
\begin{center}
  \begin{tabular}{ccc}
    \includegraphics[width=3in]{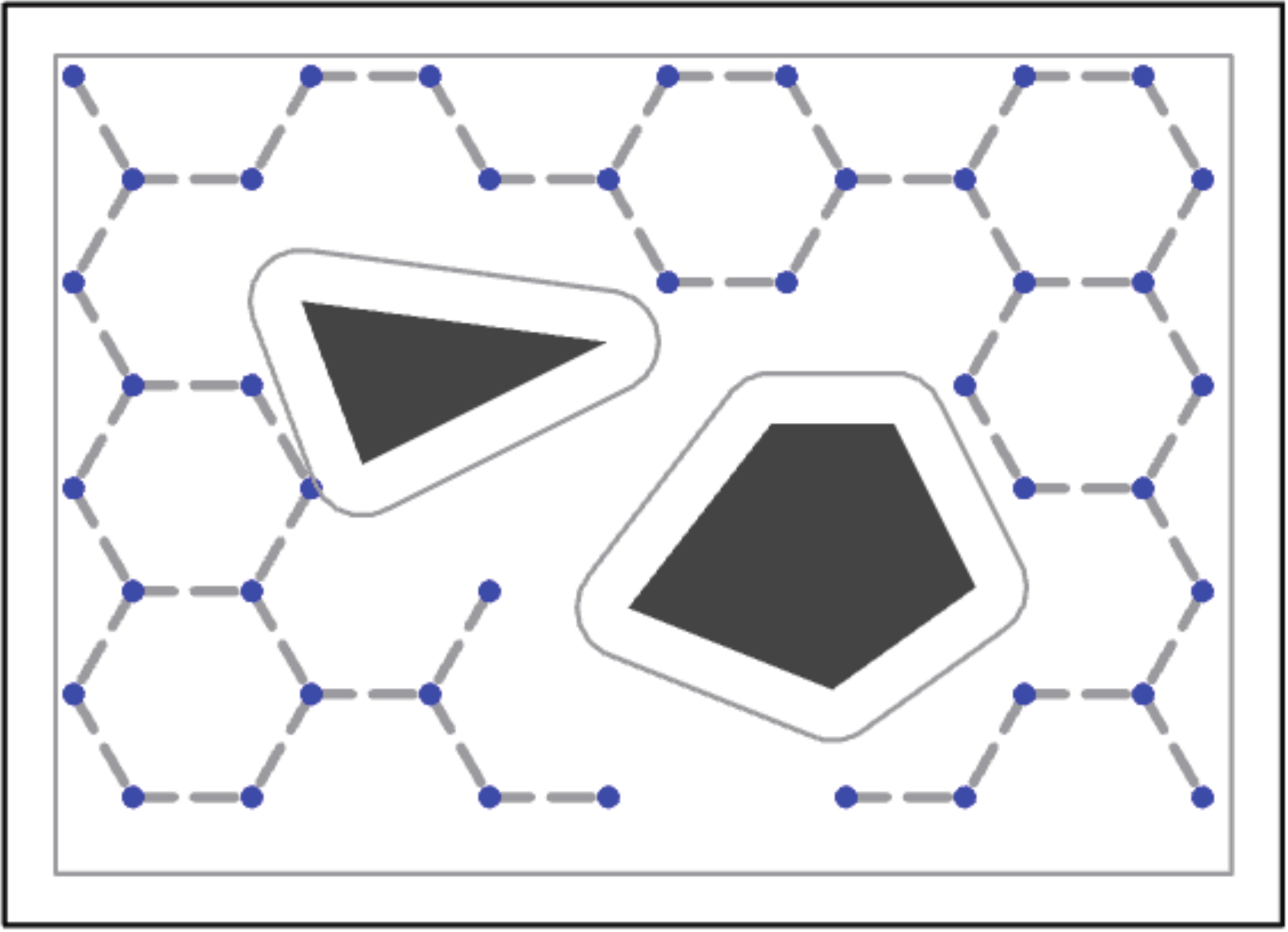} & \hspace{15mm} & 
		\includegraphics[width=3in]{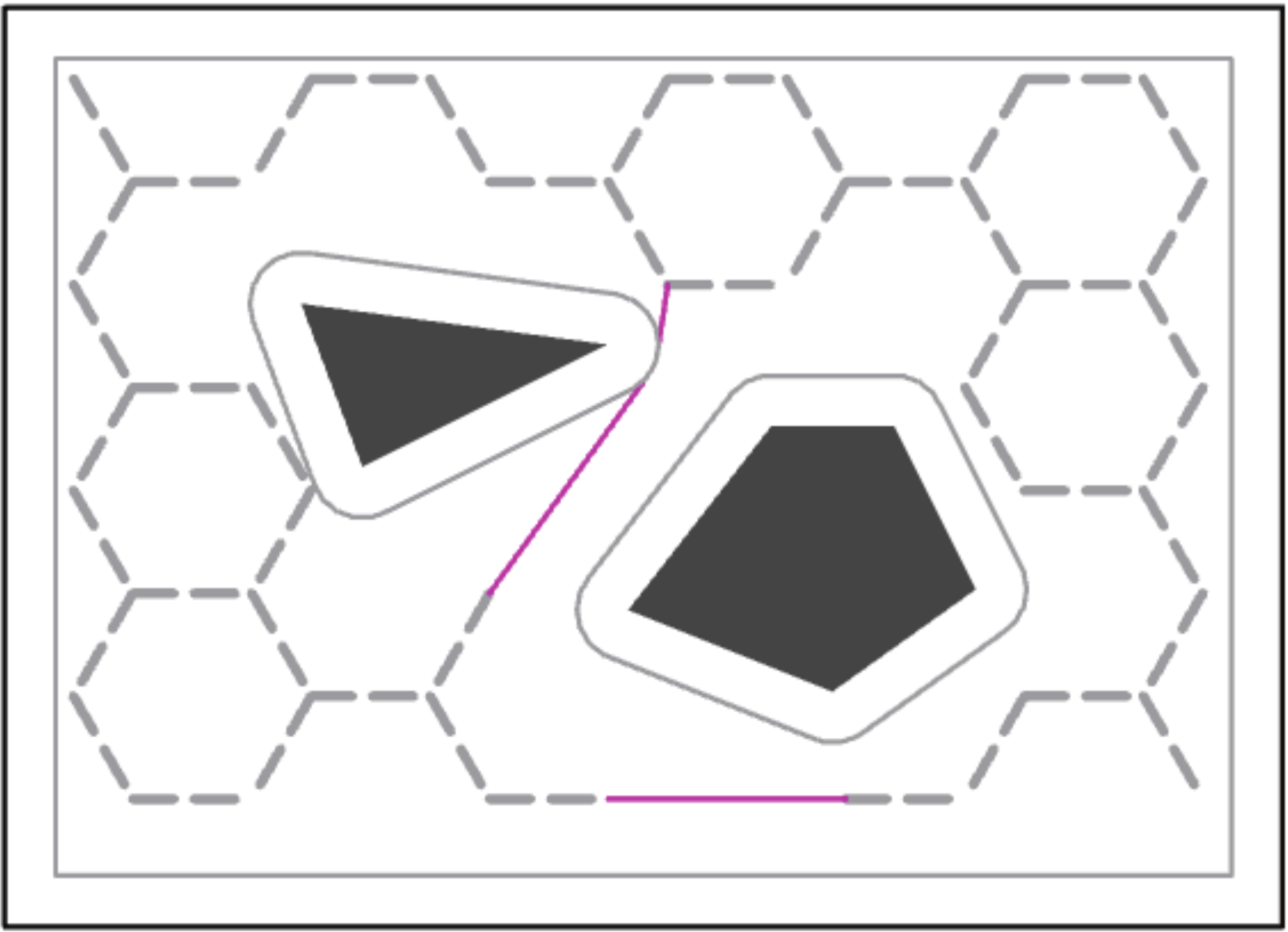}  \\ \vspace{1mm}
    (a) & & (b) \\ 
  \end{tabular}
\end{center}
\caption{\label{figure:env-example} a) An environment with a discretization that does not capture its original topology. b) The roadmap after restoring connectivity (the operations are performed automatically from our code), which then captures the topology of the original environment. }
\end{figure}

For {\em (iii)}, the discretized roadmap should capture the topology of the continuous environment well. To be more concrete, see Figure~\ref{figure:env-example}(a). In this environment, there are two holes. The lattice graph, after contraction of faces that do not contain any obstacles, does not have any holes. We expect the discrete roadmap to be connected and have number of holes (after face contraction) equal to the number of holes of the continuous environment ({\em e.g.}, Fig.~\ref{figure:example}(d)).

{\bf Remark.} We provide these assumptions only to suggest situations in which our framework is expected to perform well. In our evaluation, these assumptions are not enforced. We in fact greatly relax~\eqref{equation:separation} (from $4$ to $2.5$) and do not enforce~\eqref{equation:obstacle-separation} at all. We also give an efficient subroutine for restoring connectivity when assumption {\em (iii)} is not satisfied. For example, the routine, when applied to the example in Fig.~\ref{figure:env-example}(a), yields the result in Fig.~\ref{figure:env-example}(b), which is a screen capture from our program. We also emphasize that, given that optimal \mpp\,is an extremely challenging task computationally \cite{hopcroft1984complexity} and our focus on method effectiveness, we do not consider the problem from the angle of solution completeness.

\section{Algorithmic Framework Architecture and Roadmap-Based Discrete Problem Construction}\label{section:algorithm}
We solve the proposed problem using an algorithmic framework with two algorithmic components--discretization of the continuous problem followed by resolution of the roadmap-based problem. The overall framework contains four sequential procedures:
\begin{itemize}
\item[{\em (i)\;\;}] select and overlay a regular lattice structure over the configuration space,
\item[{\em (ii)\;}] restore environment connectivity lost in the discretization process,
\item[{\em (iii)}] snap start and goal locations to roadmap nodes to create a discrete problem on the roadmap, and 
\item[{\em (iv)}] solve the discrete \mpp\, problem optimally or near-optimally. 
\end{itemize}

We note that, when compared with motion planning methods such as PRM \cite{KavSveLatOve96} and RRT \cite{Lav98c}, our framework, looking somewhat similar on the surface, is in fact rather different. In methods like PRM and RRT, the discretization deals with the configuration space encompassing all degrees of freedom of the target system. Our approach, on the other hand, performs a careful, mostly uniform discretization of the configuration space for a single robot with two degrees of freedom. In doing so, we trade probabilistic completeness for the faster computation of near-optimal solutions. In the rest of this section, we describe the first key component of our algorithmic framework--the construction of the roadmap-based discrete problem, which subsumes the first three algorithmic procedures of the overall framework. 

\subsection{Lattice Selection and Imposition}
\noindent\textbf{Appropriate lattice structure selection} In selecting the appropriate lattice structure, we aim to allow the packing of more robots simultaneously on the resulting roadmap and obtain the structure fast. Clearly, if an insufficient number of nodes exists in the roadmap, the resulting discrete problem can be crowded with robots, which is difficult to solve and may not even have a solution. On the other hand, to allow a clean separation between the roadmap building phase and the discrete planning phase of the framework, the nodes cannot be too close to each other, {\em e.g.}, two robots occupying two different nodes should not be in collision. Moreover, it is desirable that two robots moving on different edges in parallel will not collide with each other. 

\begin{figure}[htp]
\begin{center}
  \begin{tabular}{ccc}
	\includegraphics[width=3in]{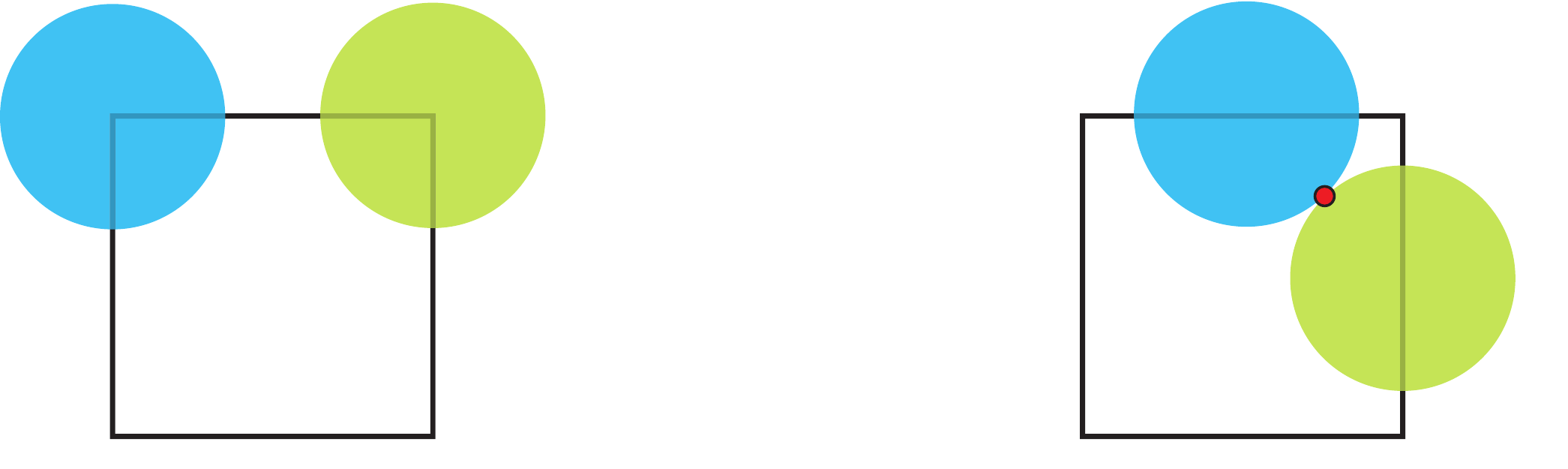} & \;\;\;\;\;\;\;\;\;\;\;\;\;\; &
	\includegraphics[width=3in]{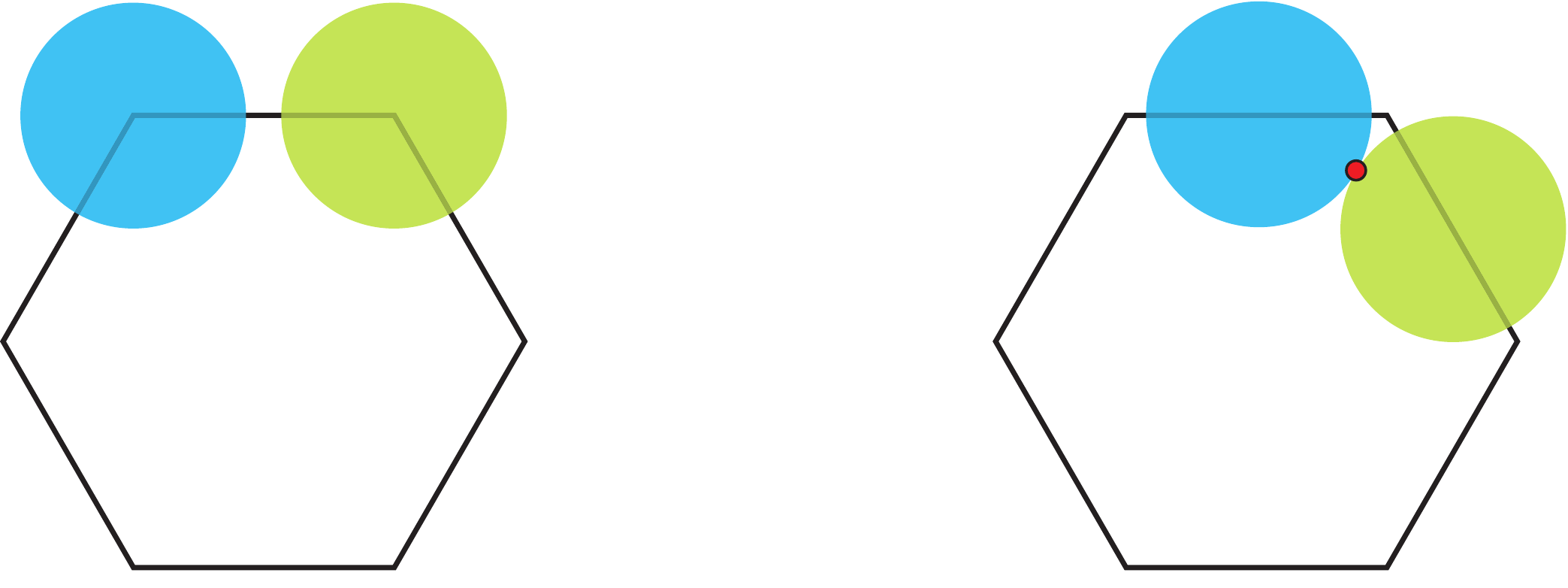}\vspace{1mm} \\
	(a) && (b) 
  \end{tabular}
\end{center}
\caption{Minimum distance between robots. To ensure no collision when executing a discrete plan, the distance between two lattice nodes must be $4/\sqrt{2} + \epsilon$ for square tilings (a) and $4/\sqrt{3} + \epsilon$ for hexagonal tilings (b). At exactly $4/\sqrt{2}$ (resp. $4/\sqrt{3}$) the robots will touch when reaching the midpoint of the edge. The contact point is shown as a red doc in both figures.} \label{fig:collision-grid}
\end{figure}

Considering all these factors together, we resort to adopting uniform tilings of the plane \cite{robert1978geometrical}. A uniform tiling of the plane is a regular network structure that can be repeated infinitely to cover the entire two-dimensional plane. Due to the regularity of uniform tilings, it is computationally easy to overlay a tiling pattern over $\mathcal C_f$. Choosing such a tiling then relieves us from selecting each node for the roadmap individually. Over the $11$ uniform tilings\footnote{These tilings are: triangular, trihexagonal, square, elongated triangular, hexagonal, truncated square, truncated trihexagonal, truncated hexagonal, snub square, rhombitrihexagonal, snub hexagonal.} of the plane \cite{robert1978geometrical}, we computed the density of robots supported by each. To allow concurrent moves of robots on nearby edges, take square tiling as an example, a square must have a side length of $4/\sqrt{2}$ to avoid potential collision incurred by such moves (see, {\em e.g.}, Fig. \ref{fig:collision-grid}(a)). Indeed, it is straightforward to show that the closest inter-robot distance is reached when two robots are in the middle of two edges connecting to the same node. For hexagonal tilings, this results in a minimum side length of $4/\sqrt{3}$ (Fig. \ref{fig:collision-grid}(b)). 

After obtaining the required side length parameters for all $11$ tilings, the maximum robot density allowed by these tilings can then be computed. We compute the density by assuming that all nodes of the regular tiling patterns are occupied by robots and compute the ratio between the area occupied by robots and the free space when it is unoccupied. For an infinite lattice with no obstacles, the hexagonal tiling is the best with about $45\%$ density, followed by the square tiling with roughly $39\%$ density. Triangular tilings have a density of only $23\%$. This leads us to choose hexagonal lattices as the base structure of the discrete roadmap. 
\begin{figure}[htp]
\begin{center}
  \includegraphics[width=3in]{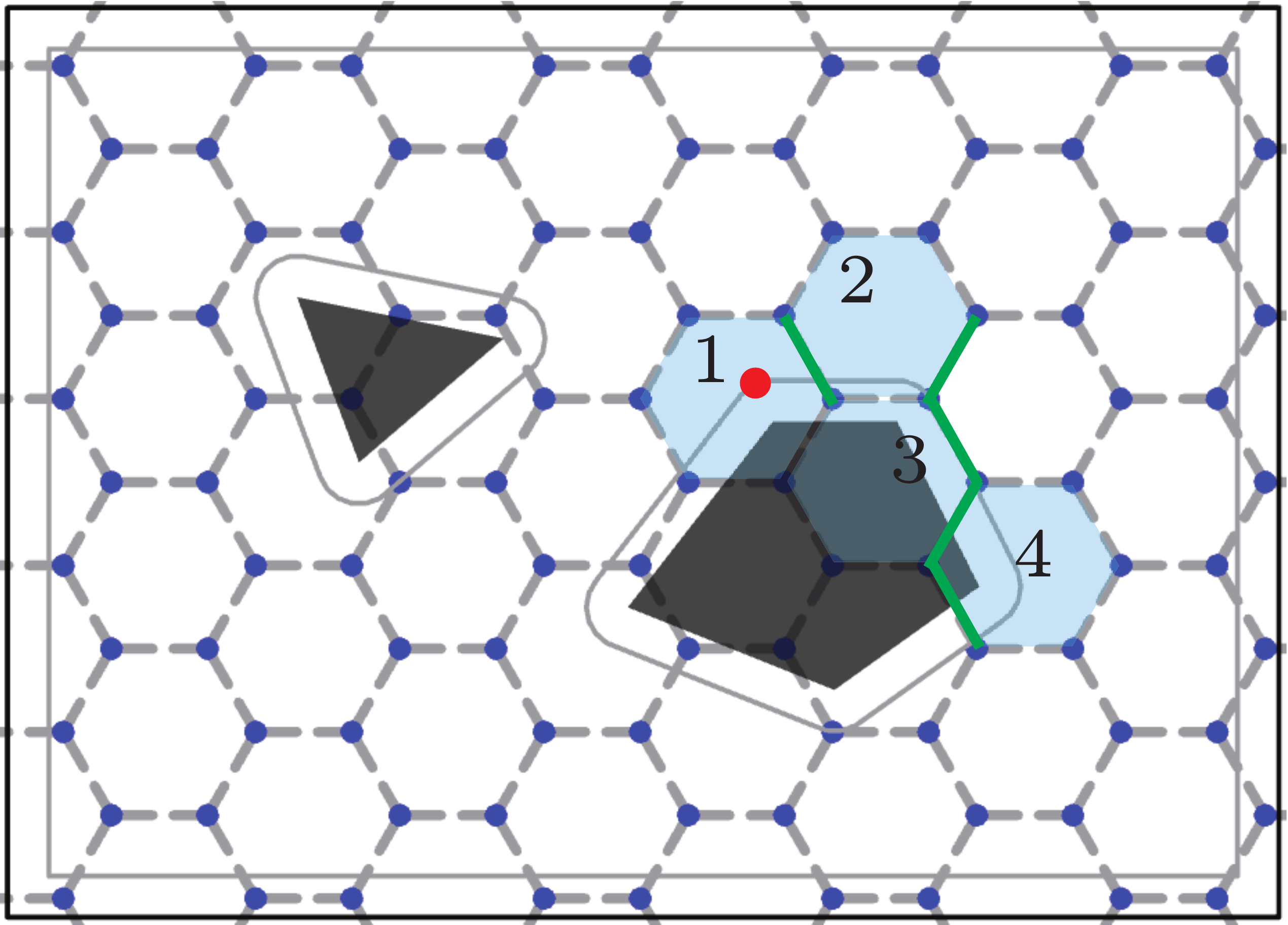}
\end{center}
\caption{Efficient computation of the hexagonal lattice that falls inside $\mathcal C_f$.} \label{fig:grid-computation}
\end{figure}

\noindent\textbf{Imposing the lattice structure } After deciding on the lattice structure, we need a procedure for imposing the structure on $\mathcal C_f$. Essentially, every edge must be checked to determine whether it is entirely contained in the free configuration space $\mathcal C_f$. Note that if this is performed naively, {\em i.e.}, performing collision checking of each edge with all obstacles, the overall complexity is on the order of $O(mA)$, in which $m$ is the complexity of the workspace and $A$ is the area contained in the outer boundary. The naive approach quickly becomes time-consuming as either $m$ or $A$ grows.

To complete this step efficiently, we start by making an arbitrary alignment between a sufficiently large piece of the infinite hexagonal lattice and the continuous environment (Fig.~\ref{fig:grid-computation}). Then, we look at one C-space obstacle (including the outer boundary) at a time. For each obstacle, we pick an arbitrary vertex on the boundary (red dot in Fig.~\ref{fig:grid-computation}) and locate the hexagon from the lattice it belongs to (in case of the example in Fig.~\ref{fig:grid-computation}, the shaded hexagonal with the label ``1'' ). We then follow the obstacle boundary and find all (green) edges of the lattice that intersect the boundary. The edges found this way do not belong to $\mathcal C_f$ and the final discrete graph structure; moreover, they partition the lattice into pieces that are either completely inside $\mathcal C_f$ or completely outside $\mathcal C_f$. This allows us to efficiently check whether the rest of the lattice edges belong to $\mathcal C_f$. To do so, we start with a vertex that is within $\mathcal C_f$ that also belongs to one of these green edges and perform a breath first search over the lattice structure, now with all the green edges deleted. All edges found this way must be long to $\mathcal C_f$. We repeat this until all vertices of the lattice that fall inside $\mathcal C_f$ are exhausted. Note that this BFS is a discrete search without performing geometric computation over real numbers, which can be done much faster than edge intersection checks. In the end, we obtain an output sensitive algorithm that typically takes time between $\Theta(\sqrt{A})$ and $\Theta(A)$, depending the total length of obstacle boundaries. In practice, using the said method, the computation time used by this step is trivial in comparison to the time it takes to do the discrete planning. 

\begin{figure}[htp]
\begin{center}
  \includegraphics[width=3in]{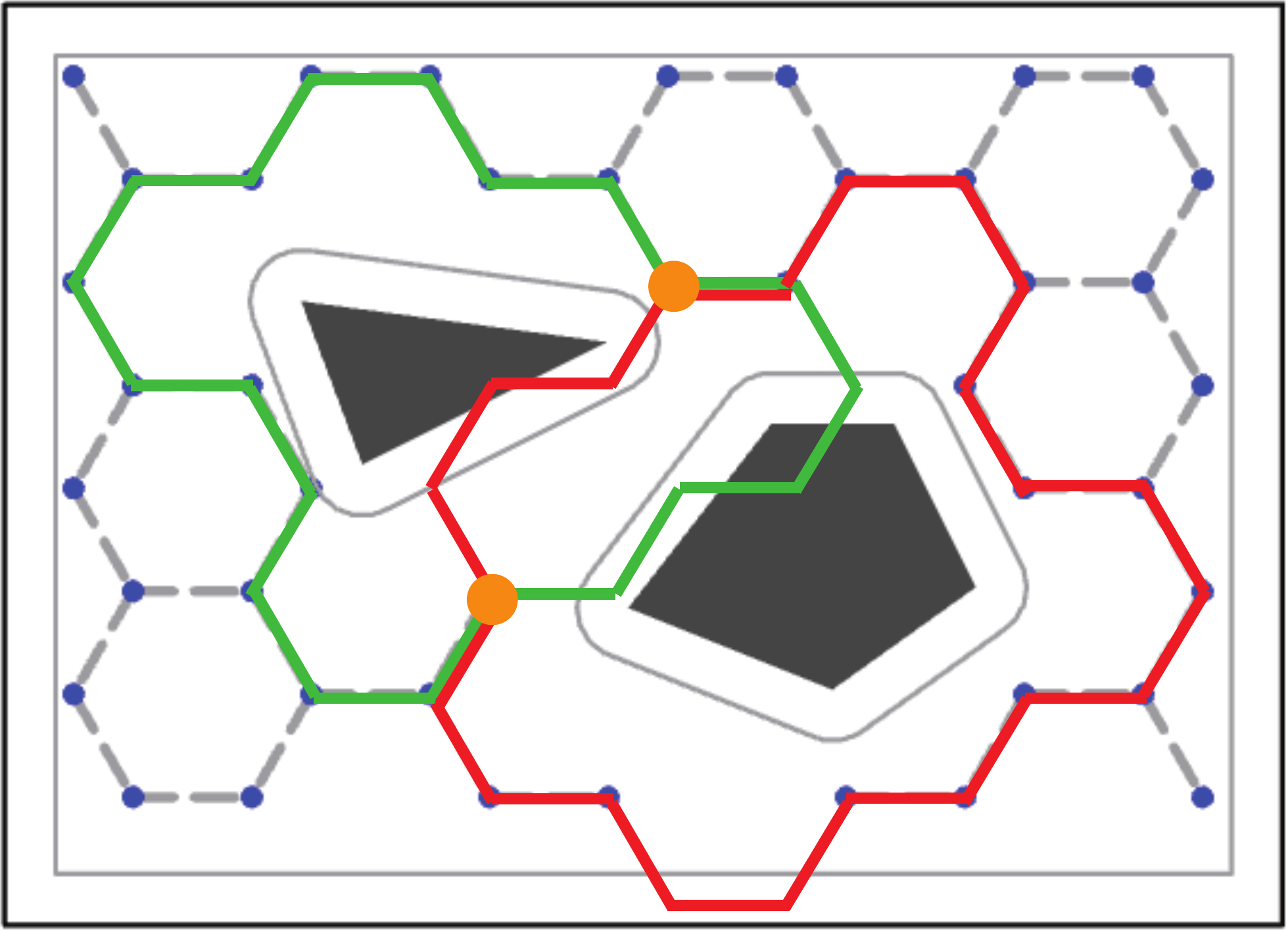}
\end{center}
\caption{Smallest cycles fully surrounding the two $\mathcal C_f$ obstacles.} \label{fig:grid-fix}
\end{figure}

\noindent\textbf{Restore Configuration Space Connectivity } We now address how we may ensure that the topology of $\mathcal C_f$ is preserved in the discrete roadmap. Essentially, we must locate places where connectivity in the continuous environment is lost. We illustrate our algorithmic solution for doing so using an example. For the problem given in Fig.~\ref{fig:collision-grid}(a), for each C-space obstacle, it is straightforward to obtain the smallest cycle on the lattice enclosing the obstacle ({\em e.g.}, the green and red cycles in Fig.~\ref{fig:grid-fix}). Then, for each pair of obstacles, we check whether the corresponding enclosing cycles share non-trivial interior and if so, locate a minimum segment on the overlapping section ({em e.g.}, the red segments between the two orange nodes in Fig.~\ref{fig:grid-fix}). Using {\em visibility graph}~\cite{lozano1979algorithm}, we may then restore the lost connectivity and obtain the roadmap shown in Fig.~\ref{fig:collision-grid}(b). Most of the computation time in this step is spent on computing the visibility graph itself, which takes time $O(m\log m + E)$ \cite{ghosh1993characterizing}, with $m$ being the complexity of the environment and $E$ being the number of edges in the resulting visibility graph.

{\bf Remark.} In the process of restoring connectivity, it is possible that the resulting roadmap cannot guarantee that simultaneous movements of disc robots are collision-free. Without getting into details, we mention that this issue can be fully addressed by sacrificing some time optimality. 

\begin{figure}[h]
\begin{center}
    \includegraphics[width=3in]{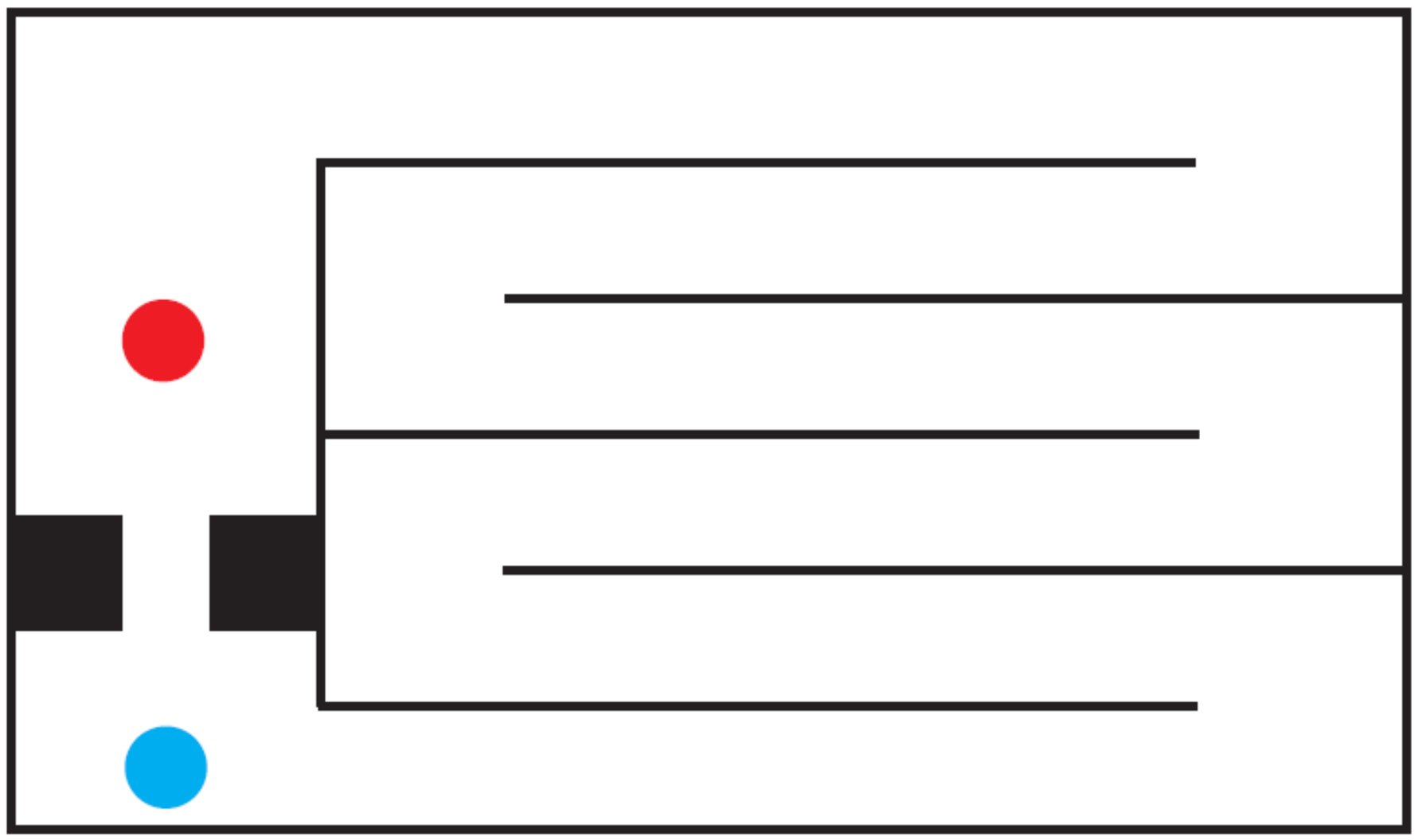}\\
\end{center}
\caption{Suppose that the start and goal locations are at the center of the blue and the red discs, respectively. If the robot does not find the narrow passage on the left, it then needs to travel through a winding path on the right. By extending the width of the environment, we can make the winding path arbitrarily long when compared to the shortest path.} \label{figure:connectivity}
\end{figure}

We also note that the preservation of the connectivity or topology of the continuous environment can be crucially important. A better connected environment has a more diverse set of candidate paths, making the resulting problem easier to solve. Perhaps more importantly, the preservation of the connectivity of $\mathcal C_f$ is essential to preserving path optimality. For a roadmap built from an overlaid square lattice, given a shortest path $p \subset \mathcal C_f$ between two points, due to the {\em strong equivalence} 
between the Euclidean metric and the Manhattan metric, the shortest path $p$ and the corresponding shortest path $p'$ on the square lattice-based roadmap are within a constant factor multiple of each other for any reasonably long path $p$ (that is, $length(p) \ll 1$ does not hold). The same argument applies to the roadmap-based hexagonal lattices. Without obstacles, the ratio $length(p')/length(p)$ over a long path $p$ is bounded by $\sqrt{2}$ for square lattices and roughly the same for hexagonal lattices. The ratio is largely the same when obstacles are present. On the other hand, if the connectivity of $\mathcal C_f$ is not preserved, then it becomes possible that $length(p')/length(p)$ is arbitrarily large. An example is given in Fig. \ref{figure:connectivity}. 

Once we establish that the roadmap preserves the near-optimality on path length, the same applies to time optimality. Given the preservation of near-optimality of individual paths, it does not directly imply that an optimal solution to the abstracted discrete problem also preserves optimality with respect to the original continuous problem, in terms of time or distance. However, our computational experiments show that this is generally the case when $\mathcal C_f$ has good connectivity. 

\subsection{Snapping Start and Goal Locations to Roadmap Nodes}
After the full roadmap is built, each start or goal location in $S \cup G$ must be associated with a nearby roadmap node. We call this process {\em snapping}. For the snapping step, for each $s_i \in S$, we simply associate $s_i$ with the closest roadmap node that $s_i$ can reach without colliding with another $s_j \in S$. The same process is performed for all $g_i \in G$ With the separation assumptions~\eqref{equation:separation} and~\eqref{equation:obstacle-separation}, this is almost always possible. In particular,~\eqref{equation:separation} implies that each hexagon from the lattice contains (roughly) at most one start and one goal location. Therefore, the number of nodes on the roadmap is at least twice the number of robots. In rare cases when conflicts do happen, we may apply the rearrangement algorithms ({\em e.g.}, \cite{SolYu15}) to perform the snapping step without incurring much penalty on time optimality. The completeness of this step is guaranteed by~\eqref{equation:separation} and~\eqref{equation:obstacle-separation}. 

With the snapping process complete, a discrete abstraction of the original continuous problem is obtained. For our example, this leads to the scenario captured in Fig.~\ref{figure:example}(c). If we are not interested in optimality, the discrete problem may be attempted using a non-optimal but polynomial time algorithm \cite{KorMilSpi84,YuRus14WAFR}. As stated in the individual subsections, the computation required in this section can be carried out using low-degree polynomial time algorithms. The relative time used for this portion is trivial as compared to the time required for solving the roadmap-based discrete problem. 

\section{Fast, Near-Optimal Discrete Path Planning}\label{section:splitting}
After a high quality roadmap is obtained with near-optimality guarantees on time and distance ({\em e.g.}, an optimality-preserving reduction from continuous space to discrete space), one may then freely choose an algorithm for finding solutions to the discrete abstraction (Fig.~\ref{figure:example}(c) in our example). Whereas an arbitrary number of globally optimal objectives can be conjured, four objectives are perhaps most natural. These four objectives minimize the maximum or the total {\em arrival time} or {\em travel distance}. Viewing from the angle of service provider ({\em e.g.}, delivery drones) and end user ({\em e.g.}, customers), minimizing the total distance or time allows the service provider to minimize energy cost or overall vehicle fleet usage. On the other hand, minimizing the maximum time or distance promises a more uniform service quality among customers. If minimizing the total arrival time or the total distance is the goal, then discrete search methods such as {\sc ID} \cite{StaKor11} can be applied. Here, we focus on the minimum {\em makespan} ({\em i.e.}, maximum arrival time or task completion time). We describe an effective method for minimizing the makespan \cite{YuLav13ICRA-A,YuLav13AAAI-LBP}, which is also a good proxy to minimizing the maximum travel distance. The method is an ILP-based one with an optimal baseline algorithm,  augmented with near-optimal heuristics to improve the computational performance.

\subsection{The Baseline, ILP Model-Based Algorithm} 
\begin{figure}[h]
\begin{center}
    \includegraphics[width=3in]{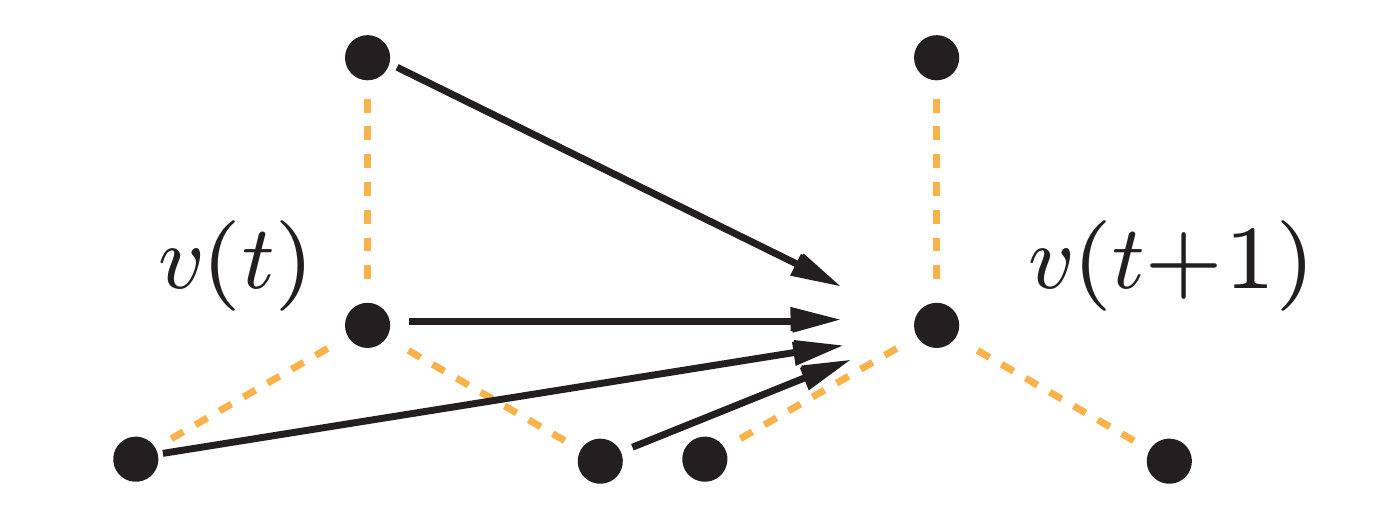} 
\end{center}
\vspace*{-3mm}
\caption{\label{figure:gadget} In a single time expansion step, a node's neighbors (including the node itself) at time step $t$ are connected to the node at time step $t + 1$.}
\end{figure}

We describe here an integer linear programming (ILP) model based algorithm from \cite{YuLav13ICRA-A}. The algorithm delivers an exact method for computing a minimum makespan solution to a discrete \mpp\, instance. The key idea is to perform {\em time expansion} over the discrete roadmap and then build the ILP model over the resulting forward only {\em space-time} graph. This allows the consideration of robot-robot time interaction to go from being implicit over the (spatial) roadmap to being explicit on the space-time graph. For a hexagonal (spatial) roadmap, between subsequent time steps, the time expansion has the intuitive local structure illustrated in Fig. \ref{figure:gadget}, which basically says that a robot at a node $v$ and its neighbors can reach $v$ in the next time step. Then, in the ILP constraint setup phase, two additional constraints are enforced:
\begin{enumerate}
\item At most a single edge leading to $v(t+1)$ from time step $t$ can be used ({\em i.e.}, all but one such edge binary variable can be set to $1$), this enforces that only a single robot may reach node $v$ at time step $t+1$. This prevents collision on a node. 
\item At most one edge from $(u(t), v(t+1))$ and $(v(t), u(t+1))$ can be used. This prevents collision on an edge. 
\end{enumerate}

The above constraints can be encoded easily using linear inequalities. To optimize for minimum time, an underestimate $T$ of the minimum required time is used as the number of time steps in the time expansion. This underestimate can be easily obtained by computing the minimum path length for each robot ignoring the rest of the robots and then taking the maximum of the path lengths. To complete the model setup, for a robot starting at node $u$ that has node $v$ as its goal, an edge $(v(T), u(0))$ is added to the model and is forced to be used. This forces a solution to find a path through the space-time graph connecting $u(0)$ and $v(T)$. If the model is infeasible, $T$ is then increased and the model is rerun until a solution is found. The number of time steps required for finding the first feasible solution is then the optimal time (for the discrete problem). When the initial roadmap has good connectivity, which is the case targeted by our work, this method appears to work reasonably well for instances with $~50$ robots, taking only minutes to solve such problems (see \cite{YuLav13ICRA-A} for details). 

\subsection{Heuristic: Divide-and-Conquer Over Time Domain}
In exploring the ILP model-based algorithm, we observe the general trend that the model solution time grows exponentially with respect to the size of the model. This prevents the baseline algorithm from being very useful as it does not work very well beyond 10-20 robots when the robot density is also high, even without the presence of static obstacles. 

\begin{figure}[htp]
\begin{center}
  \begin{tabular}{ccc}
    \includegraphics[width=1.5in]{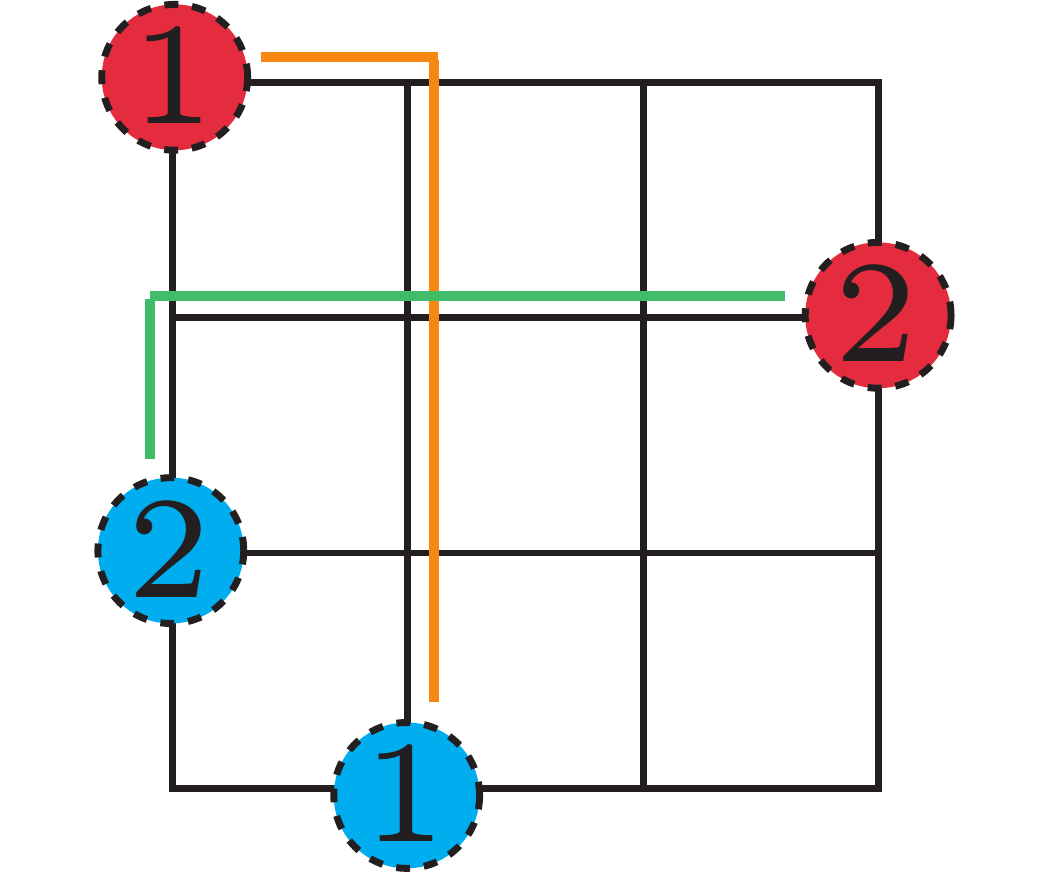} & \hspace{10mm} &
		\includegraphics[width=1.5in]{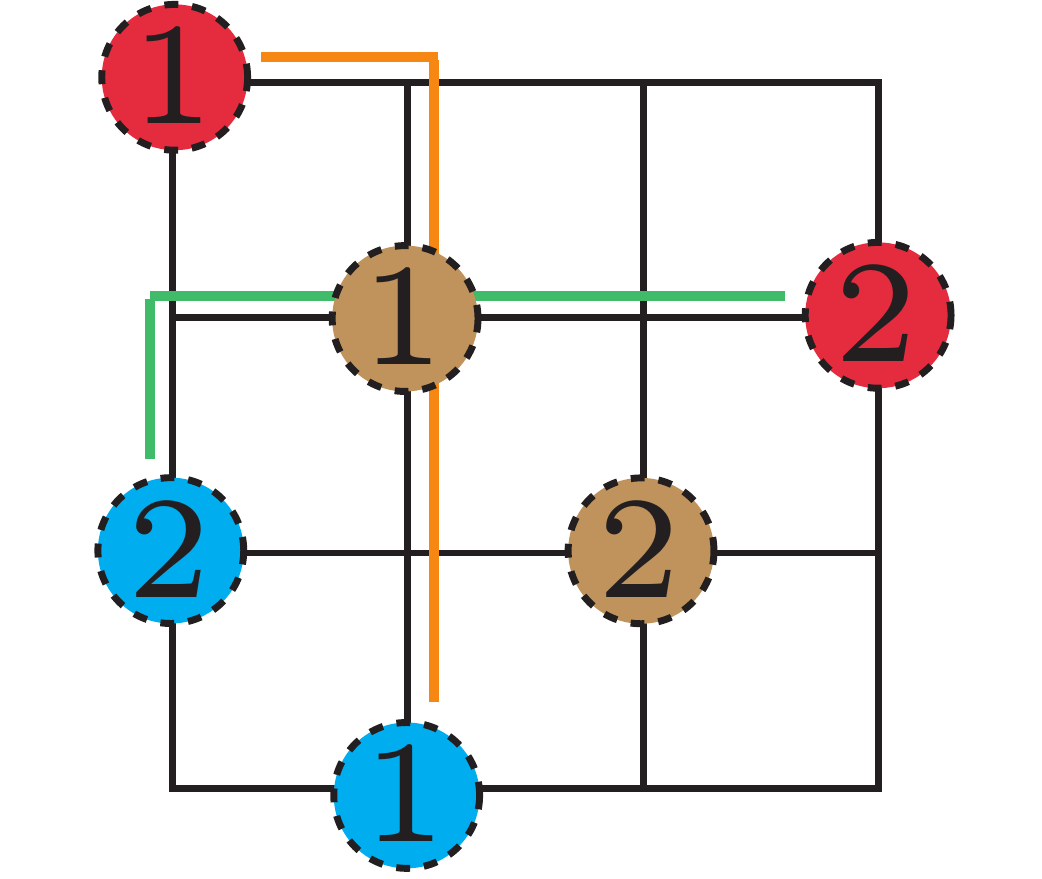}\\
		(a) & & (b)\\
  \end{tabular}
\end{center}
\vspace*{-2mm}
\caption{a) A simple two-robot problem. b) The time-divided instances.} \label{fig:dac}
\end{figure}

The same observation, though limiting the performance of the (exact) baseline algorithm, turns out to offer an useful insight toward a highly efficient divide-and-conquer heuristic. We notice that by limiting the size of the ILP model, we generally get fairly good performance from an ILP solver (we used Gurobi~\cite{gurobi} in this paper). To apply the method to more challenging problems ({\em e.g.}, solving problems with hundreds of robots quickly), we simply limit the individual ILP model that is fed to the solver. One way to achieve this is through divide-and-conquer over the time domain. We use a simple example (see Fig.~\ref{fig:dac}) to illustrate this idea. 

In Fig.~\ref{fig:dac}(a), we have a simple planning problem for two robots on a $3 \times 3$ grid. To carry out the heuristic, we first compute a shortest path for every pair of start and goal locations. In this case, we get the orange and green paths for robots $1$ and $2$, respectively. Then, if we decide to split the problem into two smaller problems, for each of the paths, it is split into two (generally) equal length pieces and the middle node is set as the intermediate goal. In our example, we may do this for robot $1$ easily and set the intermediate goal location at $(1, 1)$ from the top-left corner (the brown disc labeled $1$in Fig.~\ref{fig:dac}(b)). For robot $2$, because the middle location coincides with that of robot $1$, we pick an alternative location that is not already occupied as the intermediate goal for robot $2$, in this case $(2, 2)$ from the top-left corner. The intermediate goals for the first instance will also serve as the start locations of the second instance. This yields two child instances with both requiring a time expansion with $2$ steps each, whereas the original problem also requires a time expansion with $4$ steps. In general, we may divide a problem into arbitrarily many smaller instances in the time domain. 

If a problem is divided in this manner to $k$ sub problems, we call the resulting heuristic a {\em $k$-way split}. Because the division is over time, there is in fact no interaction between the individual, smaller instances. Once we obtain the solution for each child instance, the solutions can be glued together by simple concatenation. In practice, it turns out that this simple heuristic dramatically improves the performance without heavy negative impact on path optimality. In computational experiments, we observe a consistent speedup.

\subsection{Heuristic: Reachability Analysis}
\begin{figure}[htp]
\begin{center}
  \begin{tabular}{ccccc}
    \includegraphics[width=1.5in]{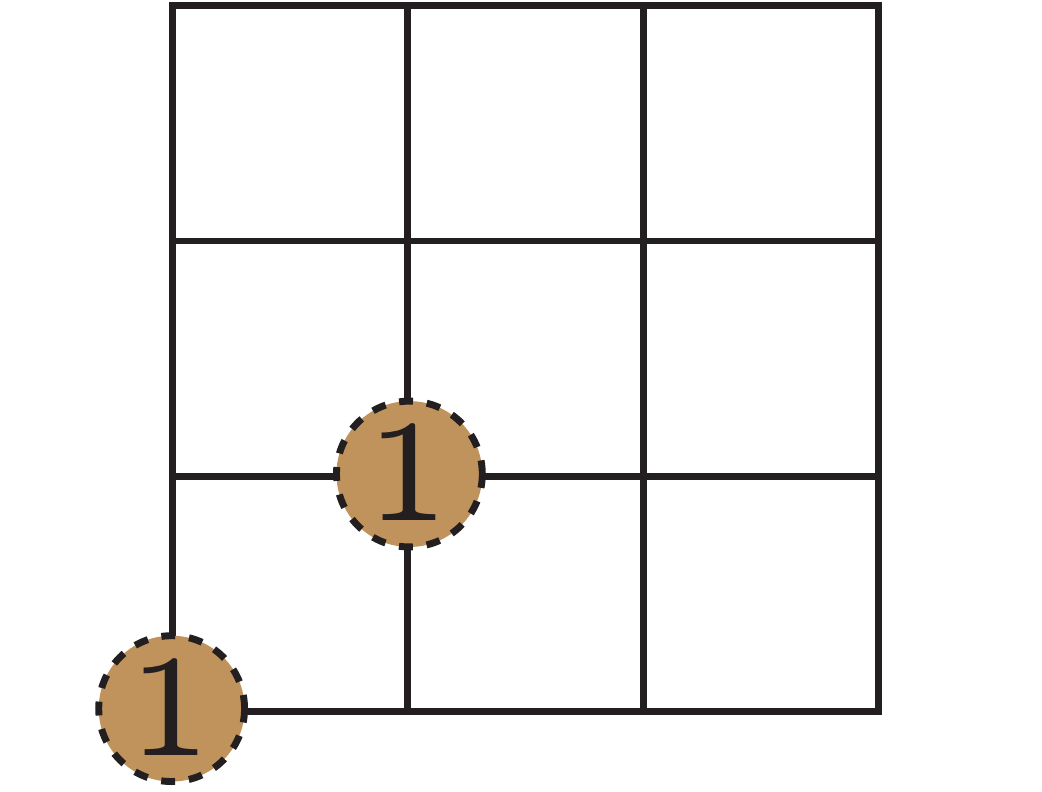} &  &
    \includegraphics[width=1.5in]{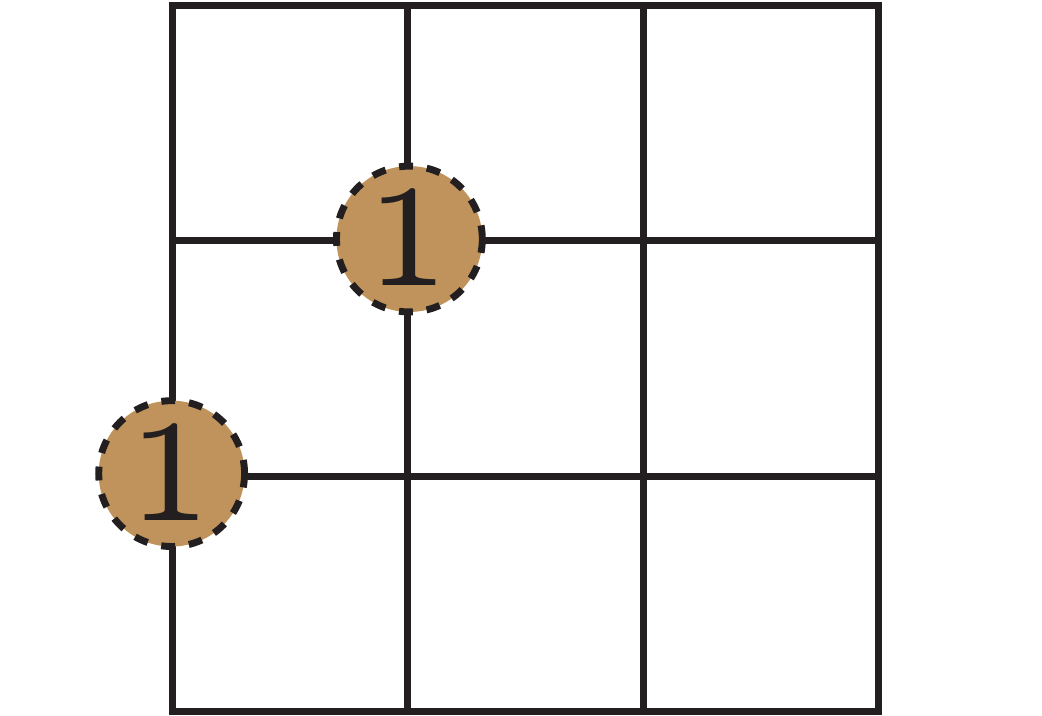} &  &
		\includegraphics[width=1.5in]{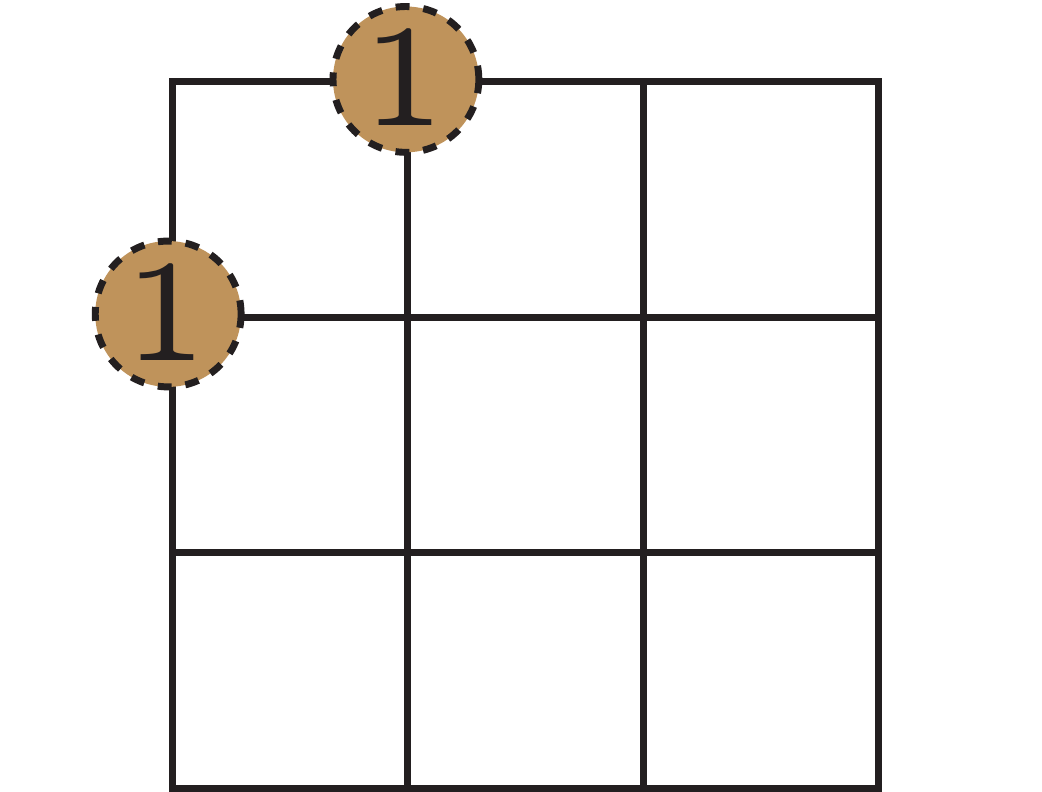}\\
		(a) & & (b) & & (c)\\
  \end{tabular}
\end{center}
\vspace*{-2mm}
\caption{The reachable portions of the $3 \times 3$ grid at time steps $t = 1, 2, 3$, respectively.} \label{fig:ra}
\end{figure}

Another method to effectively reduce ILP model size (without losing any guarantee) is through {\em reachability analysis}. Again using the example from Fig.~\ref{fig:dac}(a) and focusing on robot $1$, if the time expansion uses $4$ time steps, then the reachable nodes (from both the start and the goal) of the graph at $t = 1, 2, 3$ is illustrated in Fig.~\ref{fig:ra}. Constructing the time-expanded graph from these then greatly reduces the resulting ILP model size.  

{\bf Remark.} Because the problem we are to solve in this section is NP-complete\\ \cite{YuLav13AAAI} and we are aiming to solve it exactly, no meaningful analysis on computational complexity can be provided; we only note that the computational time required by this part of the framework dominates all other parts. 

\section{Computational Evaluation}\label{section:experiment}
\begin{figure}[ht!]
\begin{center}
  \begin{tabular}{cc}
    \includegraphics[width=3.4in]{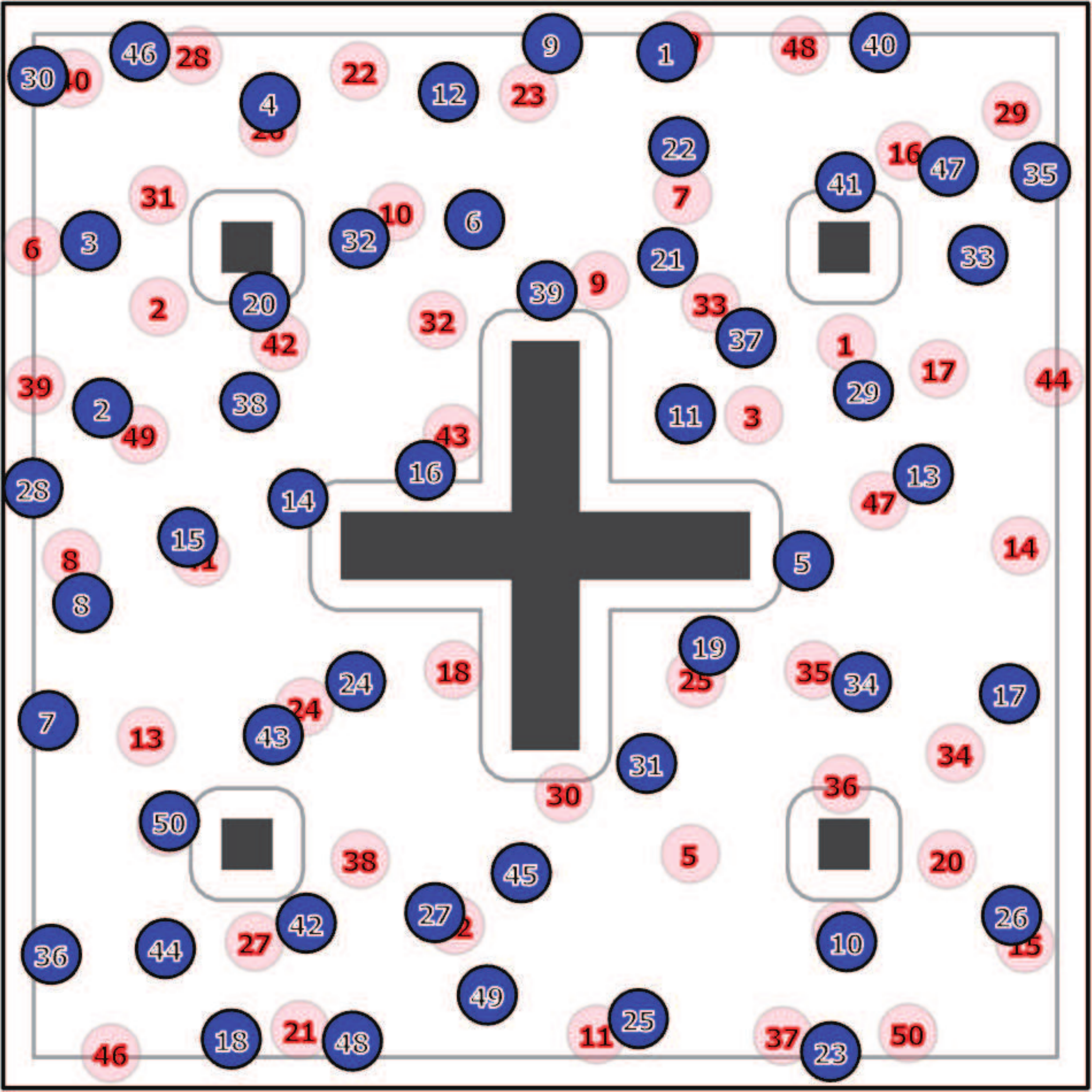} & 
		\includegraphics[width=3.4in]{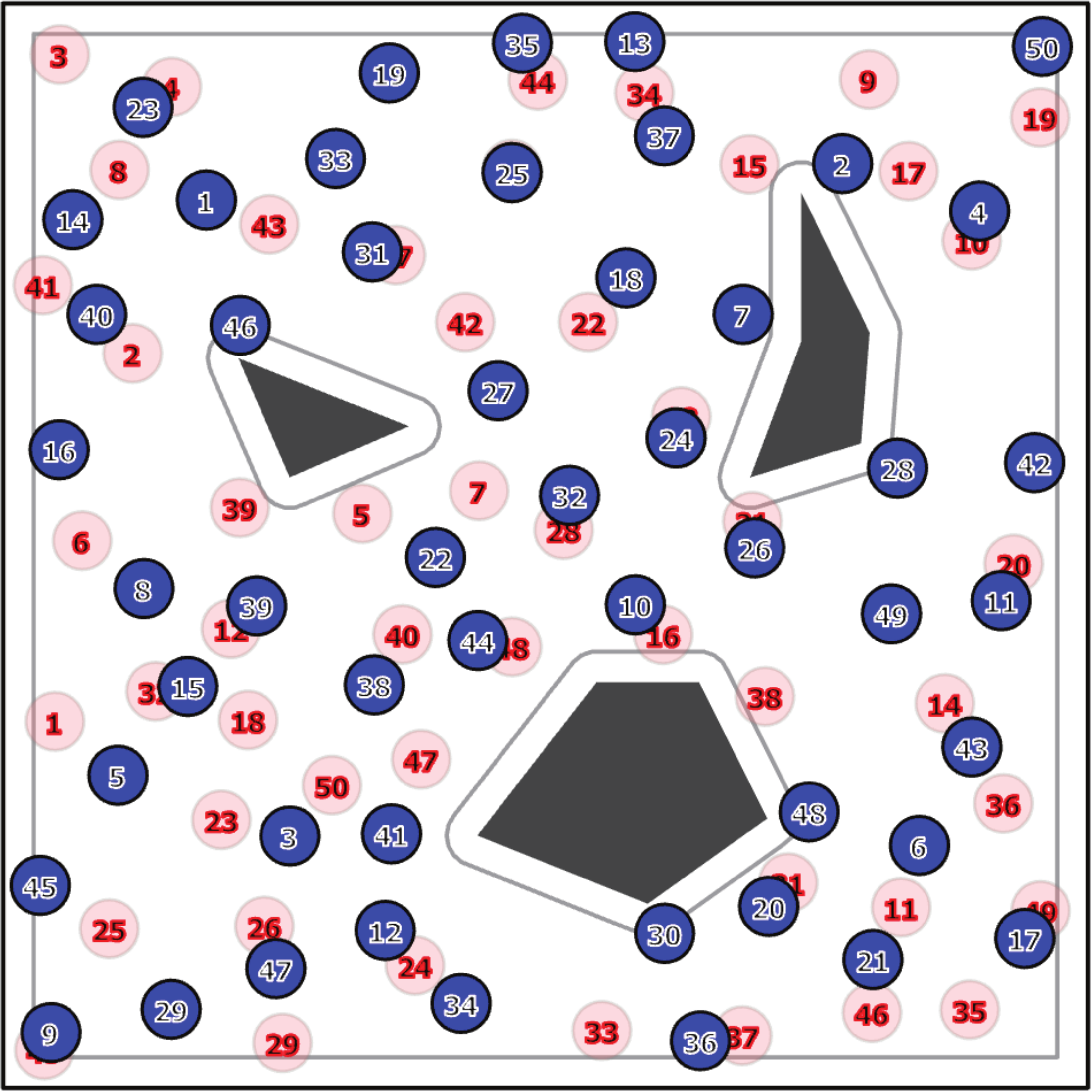} \vspace{1mm}\\ 
		(a) &  (b) \\
    \includegraphics[width=3.4in]{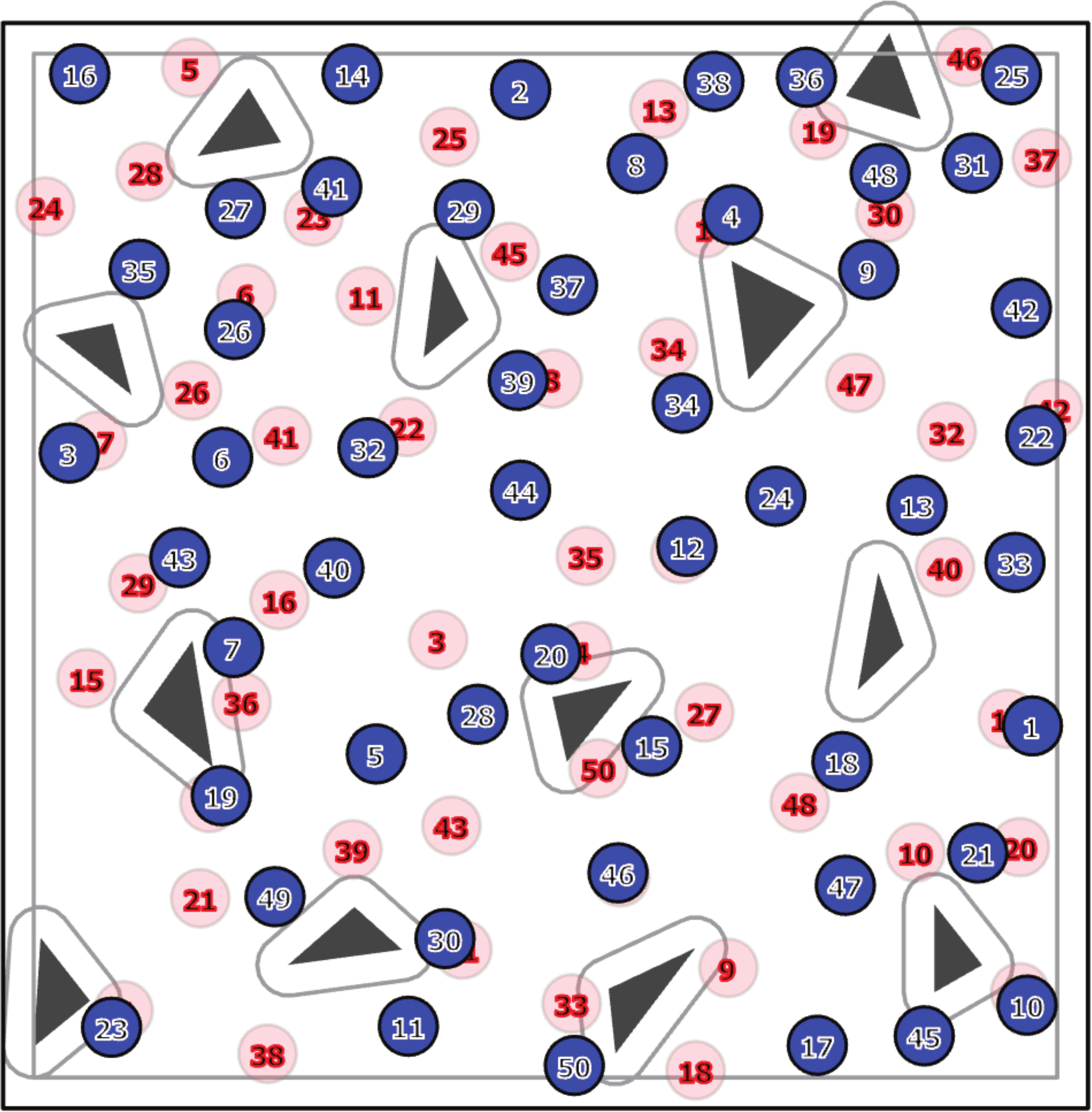} & 
		\includegraphics[width=3.4in]{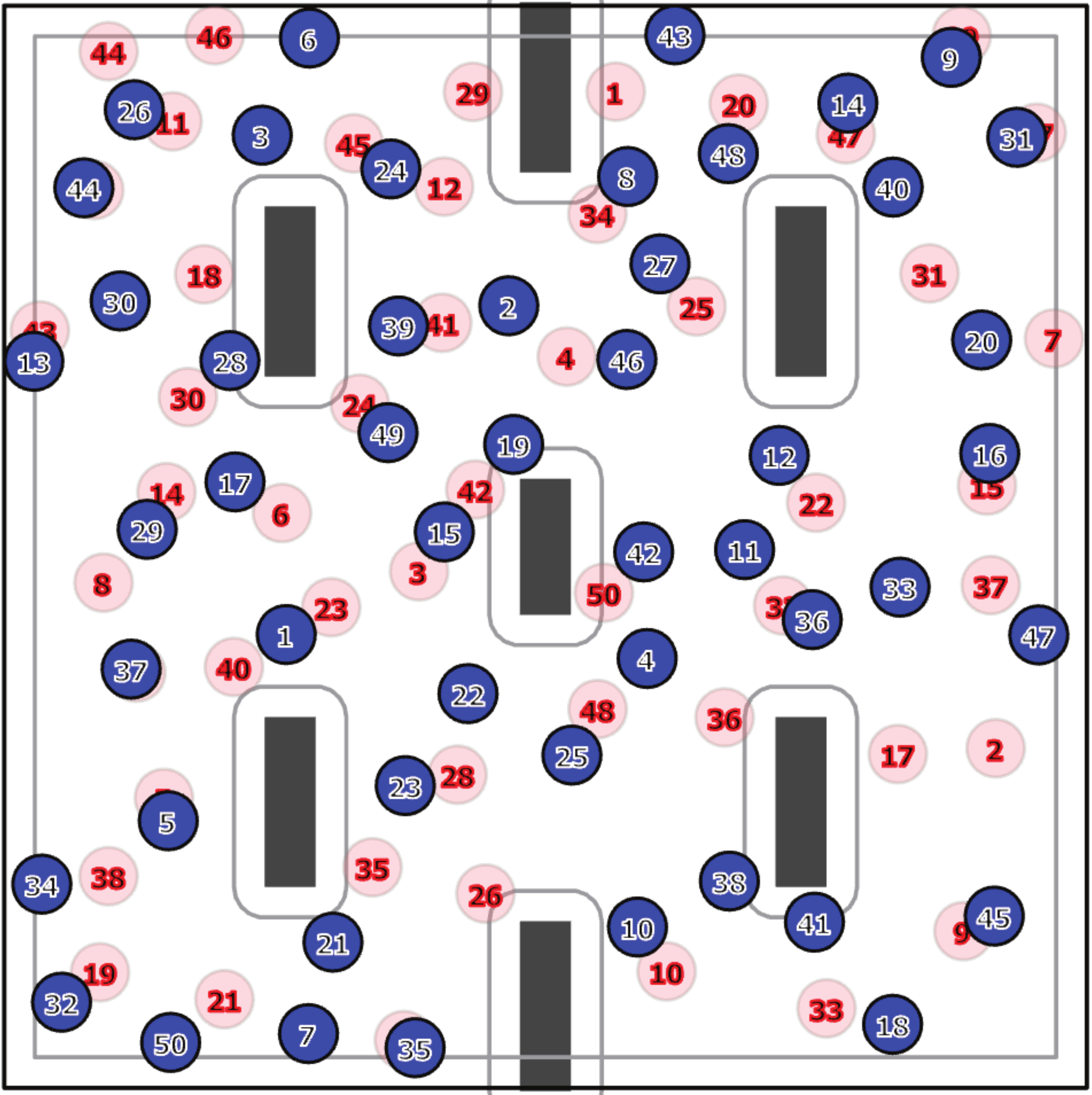}  \\ \vspace{1mm}
     (c) &(d)\\
  \end{tabular}
\end{center}
\caption{\label{figure:environments} Environments with obstacles and $50$ start and goal locations. The labeled blue discs mark the start locations and the labeled pink discs mark the goal locations. Zoom-in on the digital version of the paper for more details. a) Plus. b) (Halloween) Jack. c) Triangles. d) Bars.}
\end{figure}

We implemented the roadmap building phase in C++ using CGAL~\cite{cgal}. The discrete path planning module, written in Java, uses Gurobi~\cite{gurobi} as the ILP solver. The experiments were carried out on an Intel i7-4850HQ laptop PC. For evaluation, we tested of our algorithmic framework over five distinct environments. The first one is a simple square with a side length of 35 (recall that the robots are unit discs), with no internal obstacles. The rest of the environments have the same bounding square but contain different obstacle setups. We randomly select start and goal locations for all our tests. These environments, along with a typical $50$-robot problem instance, are illustrated in Fig.~\ref{figure:environments}. 


\subsection{Performance in Bounded, Obstacle-Free Environment}
\begin{figure}[ht!]
\begin{center}
  \begin{tabular}{ccc}
    \includegraphics[width=3in]{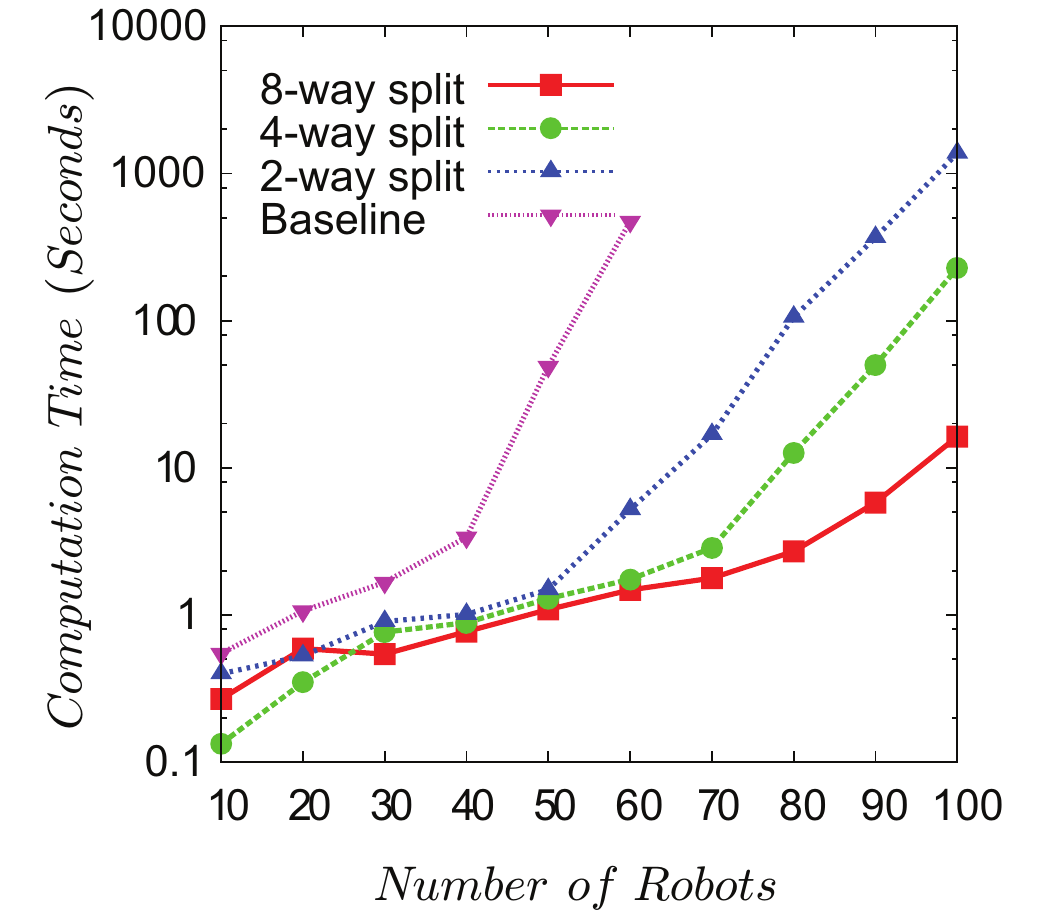} &  &
		\includegraphics[width=3in]{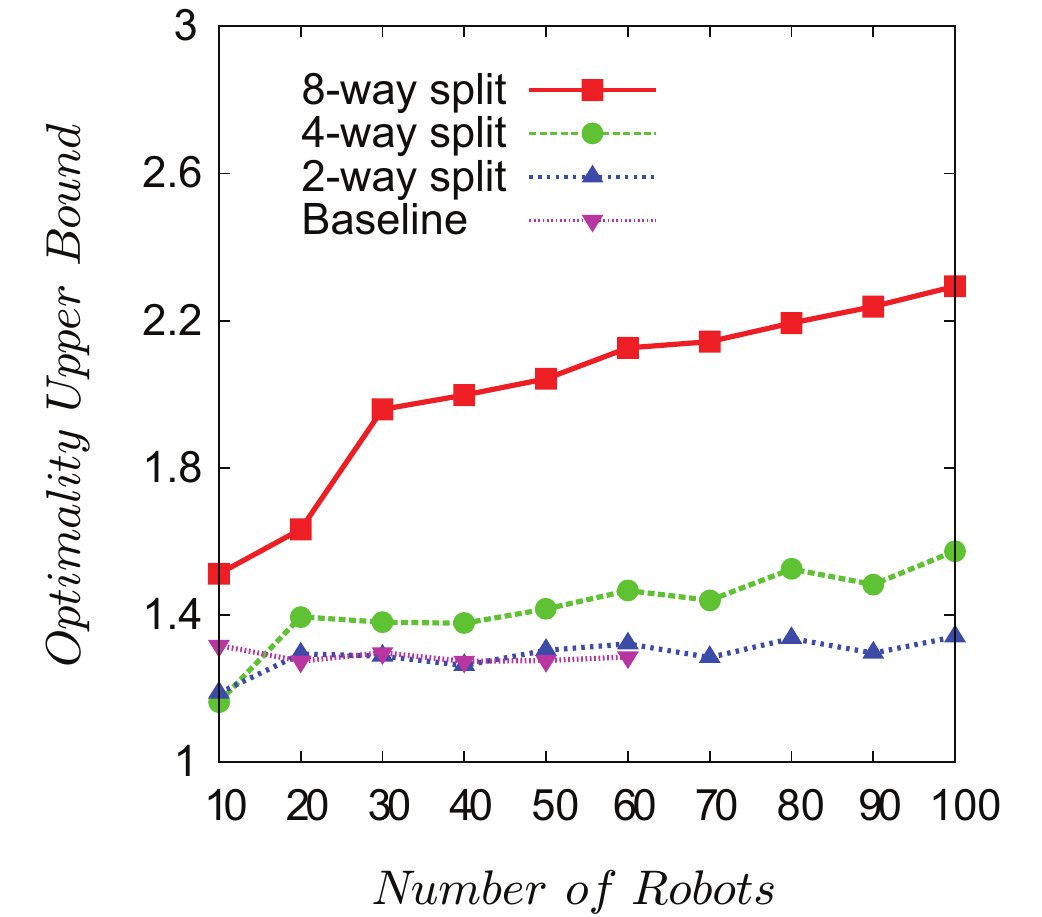}
  \end{tabular}
\end{center}
\caption{Performance of our algorithmic framework with various choices of heuristics for a square environment without internal obstacles. [left] Computation time. [right] Optimality ratio.} \label{fig:empty-eval}
\end{figure}

We first characterize how our framework performs in terms computation speed and solution optimality, as $k$-way split heuristic is used with different values of $k$. For this task, we carry out two sets of computations. The first set, covered in this subsection, focuses on bounded, obstacle-free environment. For this environment, we let the number of robots vary between $10$ to $100$ and evaluate the performance of the framework with the baseline algorithm ({\em i.e.}, a single sub-problem), 2-way split ({\em i.e.}, two sub-problems), 4-way split, and 8-way split. For each choice of the number of robots and the heuristic, $10$ test cases are randomly generated sequentially and solved. The average running time and optimality ratio is plotted in Fig.~\ref{fig:empty-eval}. Note that our computation of the optimality ratio is conservative. To compute this ratio, we find the shortest distance between each pair of start and goal locations and use the maximum of these distances as the estimate of optimal time (since the robot has maximum speed of $1$). We then obtain the optimality ratio by dividing the actual task completion time by the estimated value. 

From the experiments, we observe that the baseline algorithm actually performs quite well for up to $40$ robots in the absence of obstacles. With that said, both 2-way and 4-way splits do much better without losing much optimality--all three achieves optimality ratio between 1.2 to 1.6 in our experiments. With the 8-way split, sacrificing some optimality, we were able to consistently solve problems with $100$ robots in $10$ seconds on average. Such settings correspond to robots occupying over $25\%$ of the free space, a setting that has never been attempted before in optimal multi-robot path planning. With 8-way split, problems with $125$ robots in the same environment, which corresponds to a robot density over $31.4\%$, can be comfortably solved in about $15$ minutes. We note that, if robot density is around $20\%$, our method can readily solve problems with over $300$ robots (in a larger environment).

\subsection{Performance in Bounded Environment with Obstacles}
\begin{figure}[htp]
\begin{center}
  \begin{tabular}{ccc}
    \includegraphics[width=3in]{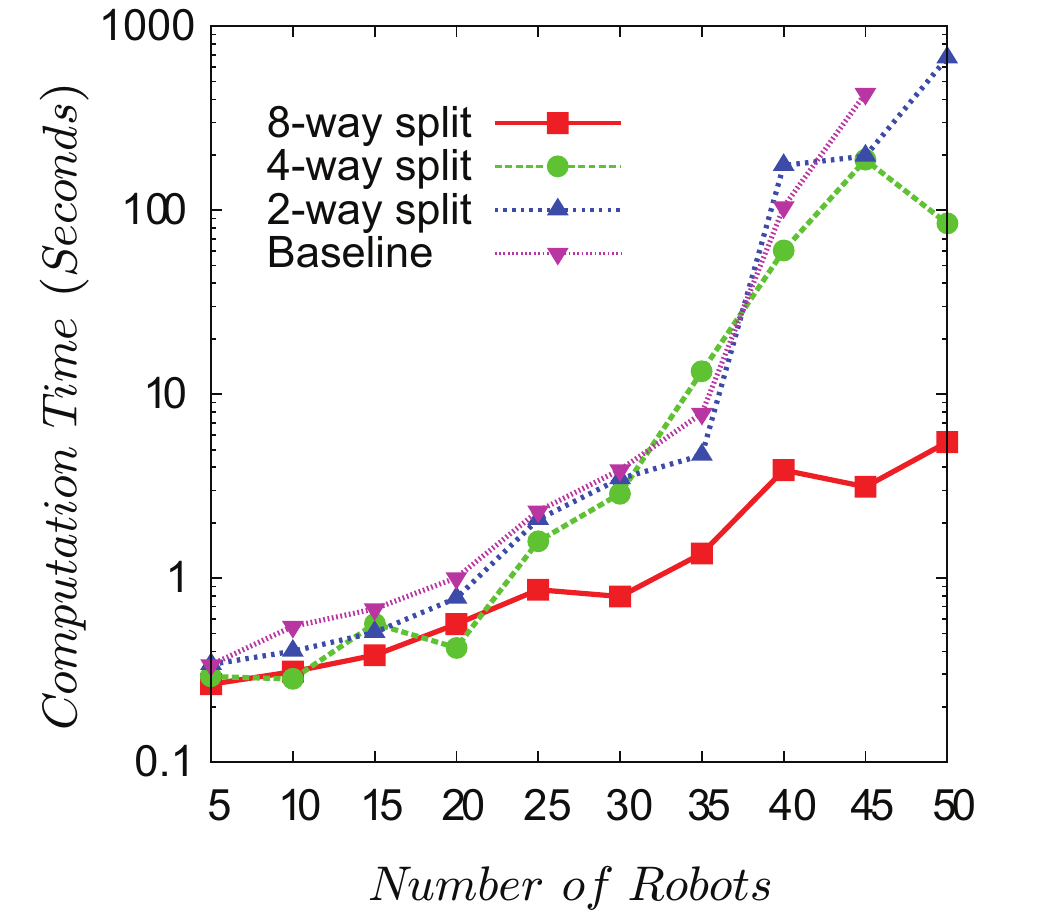} &  &
		\includegraphics[width=3in]{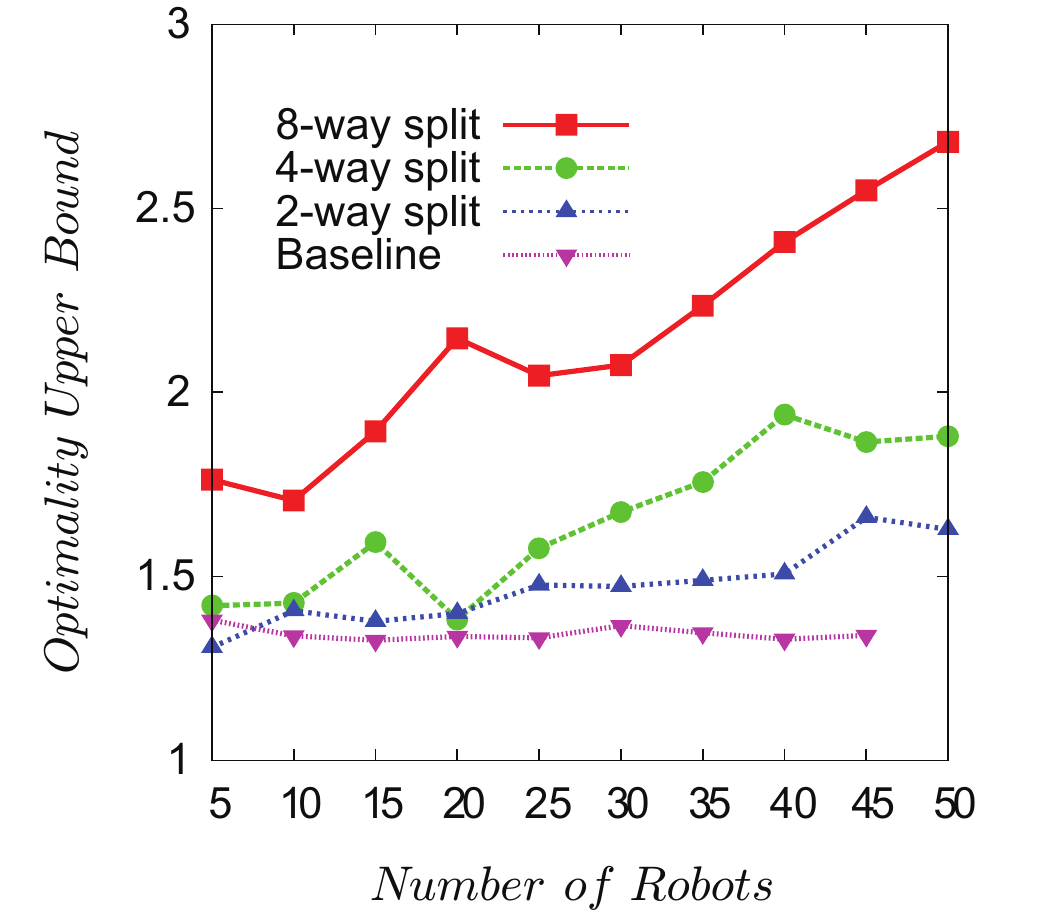}
  \end{tabular}
\end{center}
\caption{Performance of our algorithmic framework with various choices of heuristics for the ``Jack'' environment. [left] Computation time. [right] Optimality ratio.} \label{fig:jack-eval}
\end{figure}

The second set of experiments shifts the focus to an environment with obstacles. For this we use the ``Jack'' environment. We choose this environment because it is in fact a relatively difficult setting as many shortest paths have to pass through the middle, causing conflicts. The experimental result, for $5$ to $50$ robots, is plotted in Fig.~\ref{fig:jack-eval}, which is consistent with our first set of experiments. We note that obstacles, while affecting the computation time, do not heavily impact the optimality of the result.

\subsection{Evaluation of Overall Framework Performance}
\begin{figure}[htp]
\begin{center}
  \begin{tabular}{ccc}
    \includegraphics[width=3in]{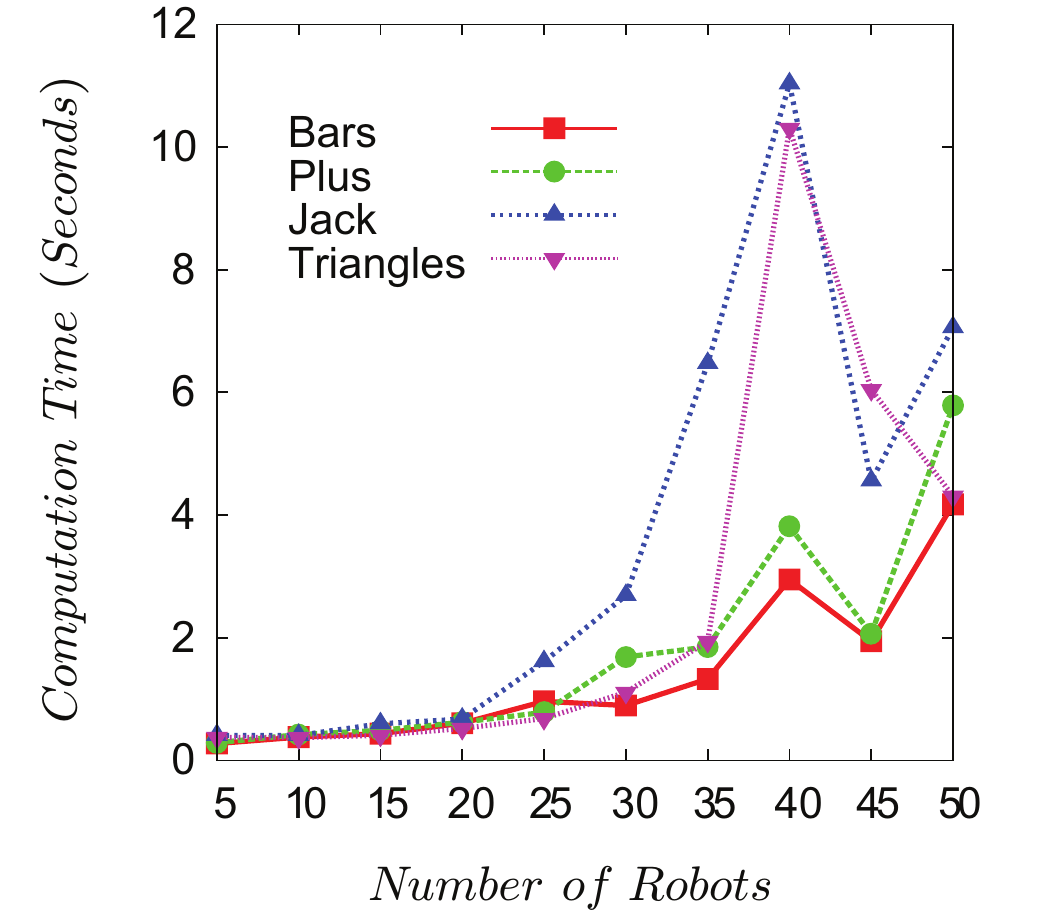} &  &
		\includegraphics[width=3in]{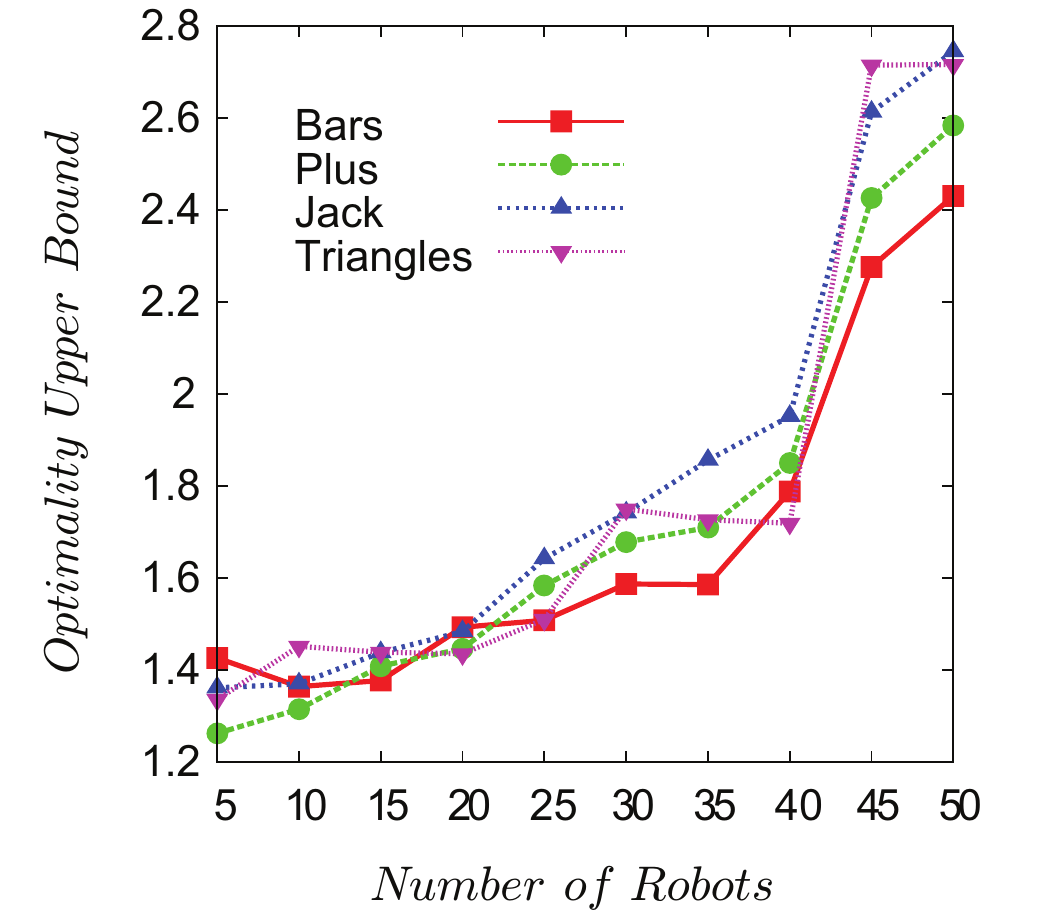}
  \end{tabular}
\end{center}
\caption{Performance of the overall framework in a wide variety of environments. [left] Computation time. [right] Optimality ratio.} \label{fig:all-eval}
\end{figure}

Our last set of experiments is aimed at showing the overall effectiveness of our framework. For this purpose we select the splitting heuristic automatically. Roughly, we do this by increasing $k$ (in a $k$-way split) to keep each time expansion with $10$ time steps, which we have found to strike a good balance between speed and optimality. For the set of environments illustrated in Fig.~\ref{figure:environments}, the experimental result is plotted in Fig.~\ref{fig:all-eval}. Our method is able to consistently solve all instances with an average solution time from $0.5$ to $10$ seconds while providing good optimality assurance on minimum makespan. The two spikes in Fig.~\ref{fig:all-eval}(a) at $40$ robots are due to the switching to 8-way split at $45$ robot for these two environments. 

\section{Conclusion}\label{section:conclusion}
In this paper, we present an algorithmic framework for tackling the multi-robot path planning problem in continuous, multiply-connected environments. Our framework partitions the planning task into two phases. In the first phase, the configuration space is tiled with a carefully selected regular lattice pattern, taking into account robot-robot collision avoidance. The imposed lattice is then processed to yield a roadmap that  preserves the connectivity of the continuous configuration space, which is essential for achieving near optimality in the final solution. Snapping the robots and their goal locations to the roadmap then transforms the initial continuous planning problem to a discrete planning problem. In the second phase, the discrete planning problem can be solved using any graph-based multi-robot path planning algorithms, after which the solution can be readily used in continuous domains. With a good optimal planner for discrete \mpp, our overall algorithm can consistently solve large problem instances with tens to hundreds of robots in seconds to minutes. 

As we make an important first step here toward a generic framework for near-optimal multi-robot path planning in continuous domains with obstacles, we also bring about many natural next steps. We discuss a few of these here, which we plan to fully explore in our future research. 

\begin{figure}[h]
\begin{center}
    \includegraphics[width=3in]{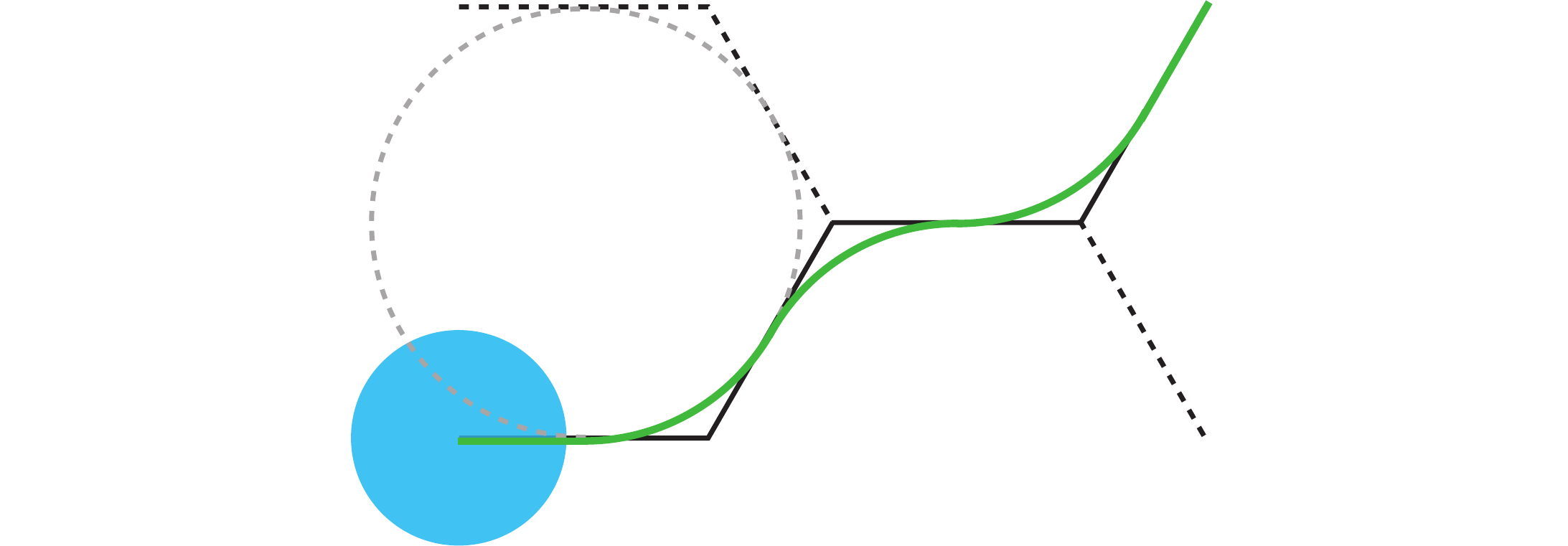}\\
\end{center}
\caption{A car-like robot with a mininum turning radius of 2 can trace any given path on a hexagonal lattice with side length $4/\sqrt{3}$ without violating its nonholonomic constraints or colliding with other robots.} \label{figure:hexagon-curvature}
\end{figure}

{\em Nonholonomic constraints.} An important issue not addressed in this paper is path planning for nonholonomic robots. We briefly touch upon this issue here. Our algorithmic framework supports quite naturally nonholonomic robots that are small-time locally controllable (STLC) with reasonable minimum turning radius. Essentially, to apply our method to a nonholonomic robot, the robot only need the capability to: \emph{(i)} move from its start location to a nearby roadmap node with a given orientation, \emph{(ii)} trace any path on the roadmap without incurring collision, and \emph{(iii)} move from a roadmap node to a nearby goal location (with an arbitrary orientation). A car-like robot, or any robot that is STLC, possesses the first and the third capabilities. Then, as long as the robot has a minimum turning radius of $2$, it can follow any path on a hexgonal lattice without violating its nonholonomic constraints (see Fig. \ref{figure:hexagon-curvature}). More importantly, multiple robots may move concurrently in such a manner without causing collisions. The introduction of nonholonomic constraints does not significantly affect optimality.

{\em Decentralized planner.} The current implementation of our framework yields a centralized algorithm. It is possible, however, to make the algorithm decentralized at the global scale. For example, we may simply let each robot perform planning individually using a method such as reciprocal velocity obstacle (RVO) based algorithm and engage locally our centralized method as the density of robots surpass some critical threshold. Note that, as the density of robots increases, RVO-based or repulsion-force-based methods generally do not have optimality guarantees and may also create deadlocks. 

{\em Optimality of hexagonal lattice in general environments.} While we have shown that a hexagonal lattice structure yields the optimal tiling in the absence of obstacles, it is unclear whether this holds well when there are obstacles in the bounded environment. In future work, we plan to study this through simulation under various obstacle settings. We will also characterize the performance using lattice structures other than hexagonal ones. The reason behind this is that, although hexagonal lattice allows the highest density, each node is only 3-connected. Square lattices, for example, has a 4-connected structure, which facilitates the discrete planning phase. Generally, discrete \mpp\, problems with higher connectivity are easier to  optimally solve.

\bibliographystyle{IEEEtranN}
\bibliography{bib}
\end{document}